\pdfoutput=1

\documentclass[11pt,a4paper]{article}

\usepackage[utf8]{inputenc}
\usepackage[T1]{fontenc}
\usepackage{lmodern}
\usepackage[margin=2.5cm,headheight=14pt]{geometry}
\usepackage{amsmath,amssymb}
\usepackage{upgreek}
\usepackage{graphicx}
\usepackage[caption=false]{subfig}
\usepackage[font=small,labelfont=bf]{caption}
\usepackage{booktabs,tabularx,multirow,array}
\usepackage{float}
\usepackage{enumitem}
\usepackage{xcolor}
\usepackage{fancyhdr}
\usepackage[numbers,sort&compress]{natbib}
\usepackage{authblk}
\usepackage{orcidlink}
\usepackage{hyperref}

\hypersetup{
  colorlinks=true,
  linkcolor=blue!60!black,
  citecolor=blue!60!black,
  urlcolor=blue!60!black
}
\newcolumntype{C}{>{\centering\arraybackslash}X}
\graphicspath{{figures/}{arxiv/figures/}}

\newcommand{\preprintnotice}{Preprint --- submitted to \textit{Applied Sciences} (MDPI), under peer review}

\fancypagestyle{plain}{%
  \fancyhf{}
  \fancyhead[L]{\small\textcolor{gray}{\preprintnotice}}
  \fancyfoot[C]{\thepage}
  }

\hypersetup{
  pdftitle={Quantifying and Mitigating Domain Shift in Peach Leaf Damage Classification: Attention Mechanisms and Fine-Tuning Strategies},
  pdfauthor={Adri\'an C\'anovas-Rodr\'iguez, Miguel A. Gonz\'alez-Ill\'an, Maria Fernanda Garc\'ia-Cruz, Pedro Nortes Tortosa, Jos\'e Salvador Rubio-Asensio, Miguel A. Zamora-Izquierdo, Juan Antonio Mart\'inez Navarro, Antonio F. Skarmeta},
  pdfsubject={Preprint submitted to Applied Sciences (MDPI)},
  pdfkeywords={peach leaf damage, artificial intelligence, computer vision, CNN, CBAM, transfer learning, plant disease}
}

\title{\textbf{Quantifying and Mitigating Domain Shift in Peach Leaf Damage Classification: Attention Mechanisms and Fine-Tuning Strategies}}

\author[1,*]{Adri\'an C\'anovas-Rodr\'iguez\,\orcidlink{0009-0002-7986-5615}}
\author[1]{Miguel A. Gonz\'alez-Ill\'an}
\author[1]{Maria Fernanda Garc\'ia-Cruz}
\author[2]{Pedro Nortes Tortosa}
\author[2]{Jos\'e Salvador Rubio-Asensio}
\author[1]{Miguel A. Zamora-Izquierdo}
\author[1]{Juan Antonio Mart\'inez Navarro}
\author[1]{Antonio F. Skarmeta}

\affil[1]{Department of Information and Communication Engineering, University of Murcia, Murcia, Spain\protect\\ \texttt{\{adriancr, ma.gonzalezillan, mariafernanda.garcia1, mzamora, juanantonio, skarmeta\}@um.es}}
\affil[2]{Department of Irrigation, Centro de Edafolog\'ia y Biolog\'ia Aplicada del Segura (CEBAS-CSIC), Murcia, Spain\protect\\ \texttt{\{panortes, jsrubio\}@cebas.csic.es}}
\affil[*]{Corresponding author: \texttt{adriancr@um.es}}

\date{}

\begin{document}

\maketitle
\thispagestyle{plain}

\begin{center}
\fcolorbox{gray}{gray!8}{\parbox{0.92\textwidth}{\small\centering
\textbf{Preprint notice.} This is a preprint of a manuscript submitted to \textit{Applied Sciences} (MDPI) and currently under peer review. It has not yet been accepted for publication, and its content may change as a result of the review process.}}
\end{center}

\begin{abstract}
Deep learning models for crop damage assessment are typically trained and validated on curated public imagery, yet their behaviour when deployed in real orchards remains poorly quantified. This work measures and mitigates that gap for peach leaf damage classification, where climate-driven abiotic and biotic stresses produce visually similar foliar symptoms. A benchmark of 1366 manually annotated peach leaves covering six damage types was assembled from public sources, and a second, independently acquired dataset of 180 field images across four classes was collected in a commercial orchard as an unseen target domain. Eleven convolutional backbones and three attention-enhanced variants were compared; CBAM-EfficientNetB5 achieved the best source-domain performance (93.3\% accuracy, 0.849 macro F1). Applied directly to the target domain, source-trained models lost on average 0.21 macro F1 points (26.5\% relative), with 12 of 14 architectures degrading, confirming that benchmark performance substantially overestimates field behaviour. Three fine-tuning strategies were then evaluated as mitigation: feature extraction proved insufficient in nearly all cases, whereas full fine-tuning recovered performance, with CBAM-EfficientNetB3 reaching 0.9459 accuracy and 0.9297 macro F1 on the local domain. Attention mechanisms improved minority-class recall and adaptation efficiency, but did not by themselves confer robustness to domain shift. The results establish a transferability baseline for peach leaf diagnosis and quantify the adaptation cost of moving from public benchmarks to operational orchards.

\end{abstract}

\paragraph{Featured application.} The transferability baseline reported in this work quantifies how much classification performance is lost when peach leaf damage models trained on public imagery are deployed in a commercial orchard, and how much of that loss is recovered with a small annotated sample from the target site. This provides growers and technical advisers with a realistic expectation of model behaviour under field conditions, and an estimate of the annotation effort required before deployment.

\section{Introduction}
\label{secIntro}

Automated image-based classification of plant damage has emerged as a key application of computer vision in precision agriculture, offering a scalable and objective alternative to manual field inspection \citep{murugavalli2025plant}. Convolutional Neural Networks (CNNs) have established themselves as the dominant approach for this task, demonstrating strong performance across a variety of crops and damage types \citep{zhao2025review,upadhyay2025deep}. However, a persistent and well-documented challenge limits real-world deployment: models trained on curated public datasets suffer significant performance degradation when applied to images captured under different field conditions, a phenomenon known as domain shift \citep{krishna2025plant,hu2025domain}. Variations in illumination, camera setup, background complexity, and phenological stage introduce distributional differences between source and target domains that standard training procedures do not adequately address \citep{shafik2025deep}.

This problem is particularly acute in perennial fruit crops, where existing datasets are small, often heterogeneous in composition, and rarely validated against locally collected field imagery. Peach leaf damage classification exemplifies these challenges: the task involves six visually similar damage categories---including bacterial spot, abiotic stress, mite presence, mechanical damage, chewing insect damage, and healthy leaves---with severe class imbalance and limited publicly available data. Prior work on this crop has focused predominantly on binary classification under controlled conditions \citep{PeachCNN,CNN_CAE}, leaving multi-class generalization across domains largely unexplored \citep{DatasetOriginalArticle,alosaimi2021peachnet,PeachEfB7}.

Deep learning, and CNNs in particular, have become the standard approach for automated plant disease detection from leaf images. Widely adopted architectures include the EfficientNet and ResNet families, DenseNet, InceptionV3, and foundational models such as VGG and AlexNet, all of which have been benchmarked across multiple crop species including tomato, potato, apple, and rice \citep{benchmarktomato,cropsCNN,RiceCNN,tomatobenchmark}. For peach foliage specifically, Yadav et al. \citep{PeachCNN} applied AlexNet, VGG16, and VGG19 to binary healthy-vs-bacterial-spot classification, while Bedi and Gole \citep{CNN_CAE} proposed a hybrid convolutional autoencoder for the same task. More recent work has combined deep learning classifiers with traditional machine learning methods such as SVM and kNN to improve robustness on multi-species datasets \citep{DLandML}. Nobel et al. \citep{DenseNetlite} introduced a streamlined DenseNet variant with gradient product optimization, demonstrating improved efficiency for tomato and apple leaf disease classification. Despite these advances, benchmarks specifically targeting multi-class peach leaf damage under realistic field conditions remain scarce.

Beyond backbone selection, attention modules have been proposed as a lightweight mechanism to improve feature discrimination without substantially increasing model complexity. The Squeeze-and-Excitation (SE) module \citep{SEpaper}, already incorporated into MobileNetV3 and EfficientNet, applies channel-wise recalibration of feature maps. The Convolutional Block Attention Module (CBAM) \citep{CBAMpaper} extends this by additionally incorporating spatial attention, enabling the model to focus on informative regions both channel-wise and spatially. CBAM has been applied in medical imaging \citep{MobileNetv2CBAM} and video surveillance \citep{EfNetCBAM}, and has been integrated into MobileNet, InceptionV3, and VGG16 for tomato leaf disease detection, yielding accuracy improvements over the corresponding baselines \citep{CBAMTomato}. Its application to multi-class peach leaf damage classification, and particularly its interaction with transfer learning under domain shift, has not been previously investigated.

A second axis of prior work concerns how models generalize once trained. Transfer learning from large-scale pretraining datasets such as ImageNet has become standard practice in plant disease classification, enabling competitive performance even with limited labeled data \citep{TLtomatoandgrave,DeepRice}. Domain Adaptation \citep{DomainAdaptation} addresses the more challenging scenario in which a distributional shift exists between the pretraining source and the deployment target, and can be described as a special case of Transfer Learning \citep{TransferLearningvsDomainAdaptation,TransferLearning}. In our context, where labels are available in the target domain, we focus on Supervised Domain Adaptation \citep{SDA}. Several strategies have been explored, including Wasserstein-based domain adaptation combined with transformer-fused convolutions \citep{tunio2024advancing}, and pipeline-based methods using metadata for plant-disease transfer \citep{cui2023plant}. Reviews consistently report that performance degrades sharply when models are applied outside their training distribution, particularly in open-field settings with variable lighting and background clutter. Supervised domain adaptation for peach leaf imagery and systematic comparisons of fine-tuning strategies under low-data regimes remain largely unexplored \citep{hu2025domain}.

Taken together, the literature reveals three persistent gaps: the absence of a comprehensive multi-class benchmark for peach leaf damage under realistic conditions; limited evaluation of attention mechanisms in this context; and, most importantly, the fact that although cross-domain degradation is widely acknowledged, its magnitude is rarely measured on independently acquired field imagery, leaving practitioners without a reference for the adaptation cost of deploying published models. This paper addresses these gaps through the following contributions:

\begin{enumerate}
\item A rigorous performance benchmark of eleven CNN architectures applied to peach leaf damage classification using a manually annotated public dataset of 1366 leaf images across six classes;
\item A systematic evaluation of the Convolutional Block Attention Module (CBAM) integrated into selected backbones, assessing its impact on minority class detection and overall classification performance;
\item A novel locally collected field dataset of 180 images acquired in a commercial orchard under real operational conditions, publicly released on Zenodo, used as an unseen target domain for evaluating cross-domain generalization;
\item A quantification of the degradation incurred by all fourteen source-trained models when applied to this target domain without adaptation, establishing an empirical transferability baseline for peach leaf diagnosis;
\item A comparative analysis of three transfer learning fine-tuning strategies---feature extraction, partial fine-tuning, and full fine-tuning---applied as mitigation, including their differential effect on attention-enhanced architectures.
\end{enumerate}

\section{Materials and Methods}

\subsection{Data Acquisition}

The data used in this project can be divided into two types: Public Datasets and Local Dataset.

\subsubsection{Public Dataset}

The public datasets were obtained from two sources: the PlantDoc dataset \citep{PlantDocDataset} and a dataset originally published alongside the article in \citep{DatasetOriginalArticle}.

PlantDoc is an open-source dataset containing 2598 images across 13 plant species and up to 17 disease classes. For peach leaves specifically, it includes 103 images. Given the variability and the relatively low quality of some samples, we manually selected only those of sufficient quality for our study.

The second dataset was retrieved from Mendeley Data, where the authors of the aforementioned article made it publicly available \citep{DatasetUsed}. This dataset contains images of peach fruits, leaves and stems, categorized into six different classes. However, these labels refer to entire images rather than individual leaves, which limits their usefulness for our objectives. Our focus is on leaf-level classification, with particular interest in the health status of each leaf rather than a mixture of plant organs. Despite this limitation, the dataset was valuable because of the conditions under which it was collected: images were captured with smartphone cameras under diverse lighting and environmental scenarios. This variety made the dataset especially suitable for studying challenging cases and compensating for the lack of open-field data. The dataset contained 1033 images in total. Since our analysis was restricted to peach leaves, we discarded images that did not clearly display leaves and retained only those relevant to our target classes.

After this selection process, we obtained 457 images containing multiple leaves. Each leaf was manually annotated using the CVAT image annotation tool \citep{CVAT}, producing polygonal annotations in COCO format. These annotations were then used to crop individual leaves and assign them to one of six classes:

\begin{itemize}
\item \textit{Bacterial spot:} Caused by \textit{Xanthomonas arboricola pv. pruni}, affected leaves show small angular spots with yellow halos and, in severe cases, central perforations and a brown-to-green color gradient at leaf tips \citep{luo2022identification,garita2018xanthomonas}. Early detection is critical to prevent defoliation and economic losses \citep{bacterial_treatment}.
\item \textit{Abiotic stress:} Leaves exhibiting symptoms of environmental stress (solar radiation, drought, or nutrient deficiencies), visually expressed as chlorosis, diffuse discoloration, or necrotic areas \citep{abiotic_visual_detection}.
\item \textit{Mechanical stress:} Leaves physically damaged by wind or cultural practices (pruning, thinning), presenting as scratches, irregular holes, or torn tissue. Despite being abiotic in origin, its distinct visual appearance justifies treating it as a separate class from abiotic stress.
\item \textit{Mite presence:} Leaves infested by spider mites (e.g., \textit{Panonychus ulmi}), showing fine webbing, white stippling, and progressive brown discoloration (``bronzing'') caused by chlorophyll extraction during feeding \citep{joshi2023management}. Detection is complicated by the pest's preference for the leaf underside \citep{mite_underside}.
\item \textit{Healthy:} Leaves showing no visible symptoms of biotic or abiotic damage, with uniform green coloration, intact margins and no chlorotic, necrotic or perforated tissue. This class acts as the control category and is the most represented one in the dataset.
\item \textit{Chewing insects:} Leaves damaged by mandibulate insects (caterpillars, beetles, grasshoppers), characterized by rounded or symmetrical holes near the leaf margins \citep{Mahendiran2022}. Severe infestations reduce photosynthetic area and can cause significant yield losses.
\end{itemize}

In total, 1366 individual leaves were annotated and labeled, resulting in 1366 cropped leaf images. The final class distribution is shown in Table~\ref{tab:class_distribution}. The dataset is unbalanced, dominated by the Healthy class. To address this, we implement data augmentation to increase the number of images in minority classes (see Section~\ref{Data_Augmentation}) and include class weights to enhance the loss function during training (see Section~\ref{Experimental_Setup}).

\begin{table}[H]
\centering
\caption{Class distribution of annotated leaf images from public databases.\label{tab:class_distribution}}
\begin{tabularx}{0.6\textwidth}{lC}
\toprule
\textbf{Class} & \textbf{Number of Images}\\
\midrule
bacterial\_spot & 144\\
mechanical\_stress & 58\\
abiotic\_stress & 118\\
mite\_presence & 50\\
healthy & 951\\
chewing\_insect & 45\\
\midrule
\textbf{Total} & \textbf{1366}\\
\bottomrule
\end{tabularx}
\end{table}

\begin{figure}[H]
\centering
\subfloat[Healthy leaf]{\includegraphics[width=4.5cm,height=4cm,keepaspectratio]{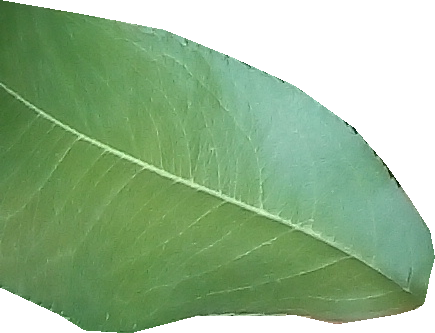}}
\hfill
\subfloat[Leaf with bacterial spot]{\includegraphics[width=4.5cm,height=4cm,keepaspectratio]{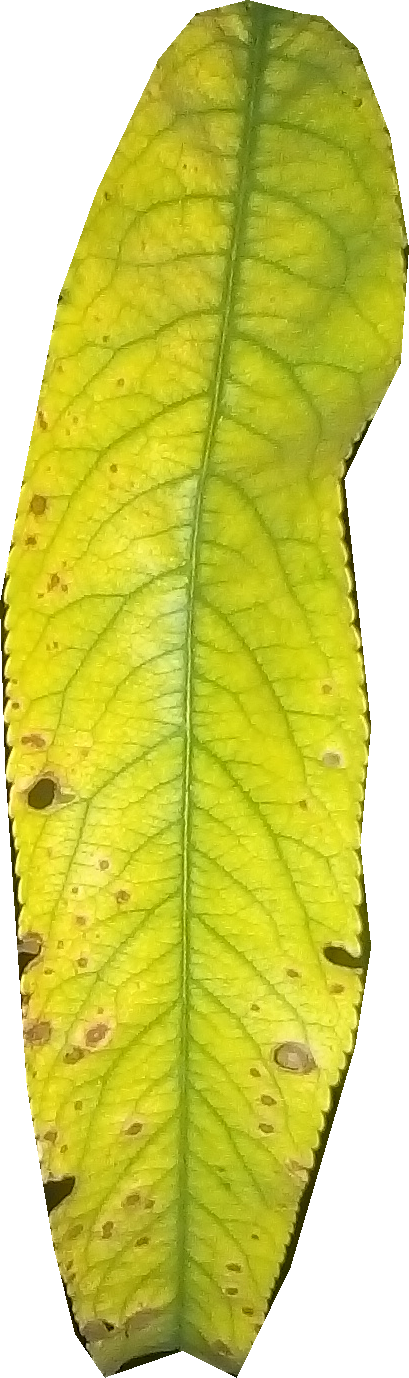}}
\hfill
\subfloat[Abiotic stress (iron deficiency)]{\includegraphics[width=4.5cm,height=4cm,keepaspectratio]{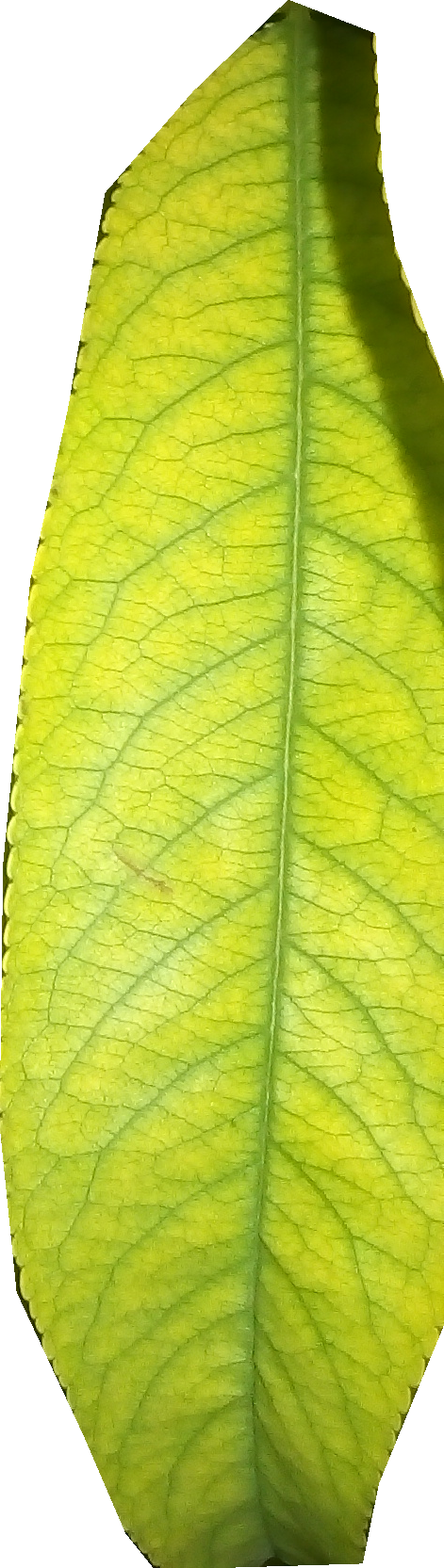}}\\
\subfloat[Mechanical stress (tearing)]{\includegraphics[width=4.5cm,height=4cm,keepaspectratio]{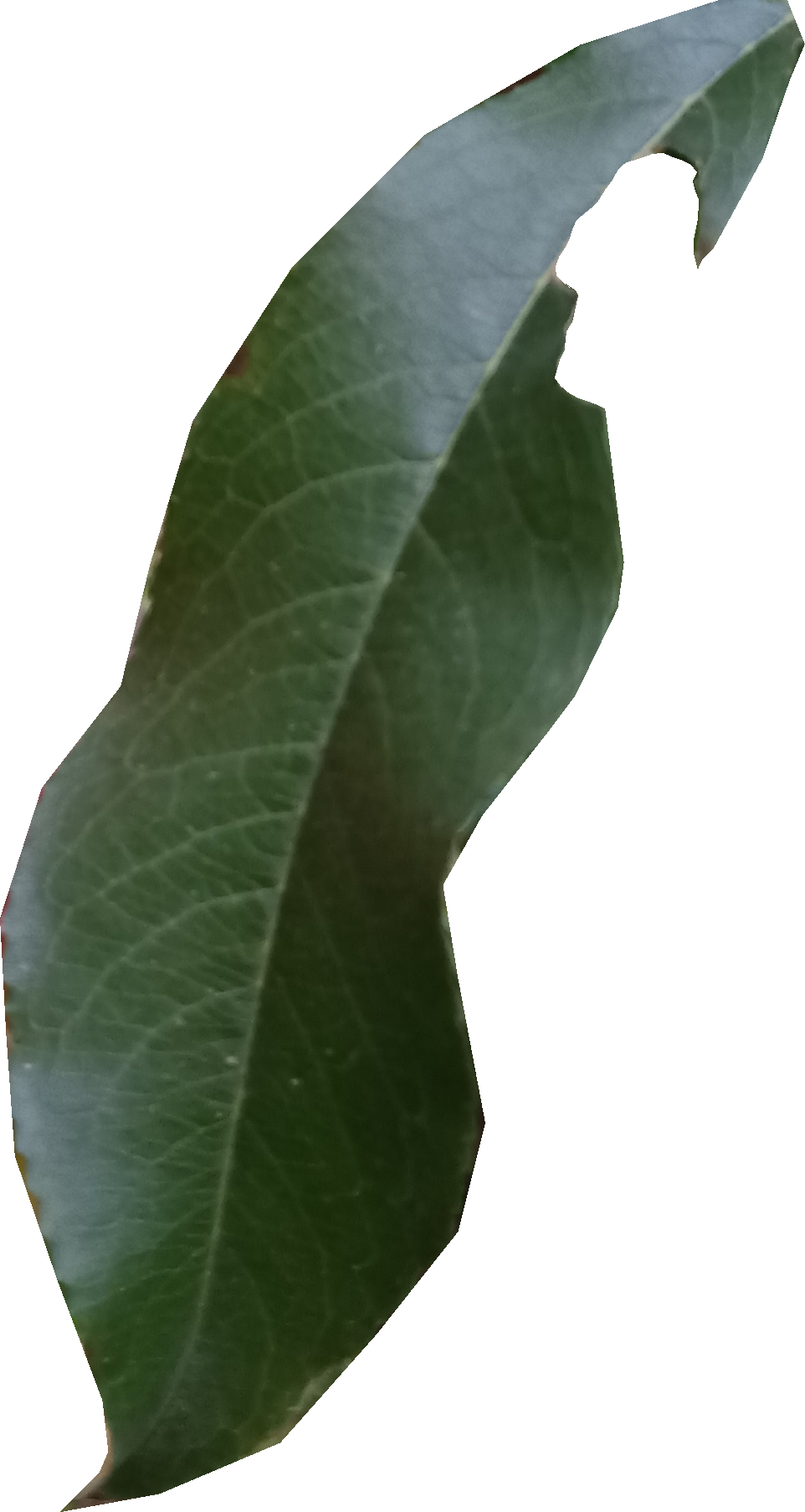}}
\hfill
\subfloat[Chewing insect damage]{\includegraphics[width=4.5cm,height=4cm,keepaspectratio]{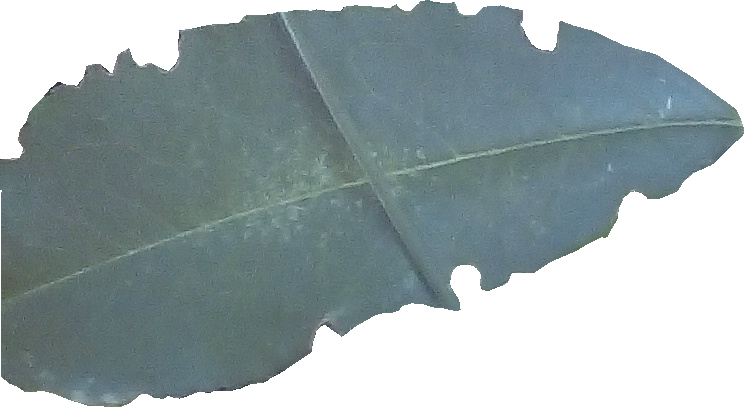}}
\hfill
\subfloat[Mite infection]{\includegraphics[width=4.5cm,height=4cm,keepaspectratio]{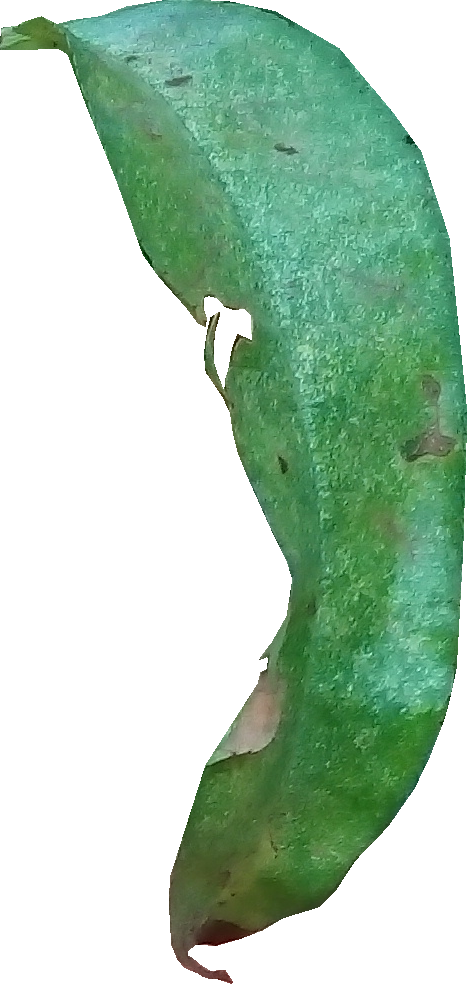}}
\caption{Representative examples of each class in the dataset: (\textbf{a}) healthy leaf; (\textbf{b}) leaf with bacterial spot; (\textbf{c}) abiotic stress (iron deficiency); (\textbf{d}) mechanical stress (tearing); (\textbf{e}) chewing insect damage; (\textbf{f}) mite infection.\label{fig:clases_dataset}}
\end{figure}

Due to the fact that the dataset was obtained from a public source, we lack information about cultivation conditions, such as the variety, the seasons in which the images were taken, or the phenological stages. This limitation justifies the need to incorporate additional local data, for which more detailed information is available, allowing us to better validate the study. In Figure~\ref{fig:clases_dataset}, a representative example of each described class is presented.

\subsubsection{Local Dataset}

\begin{figure}[H]
\centering
\includegraphics[width=10cm]{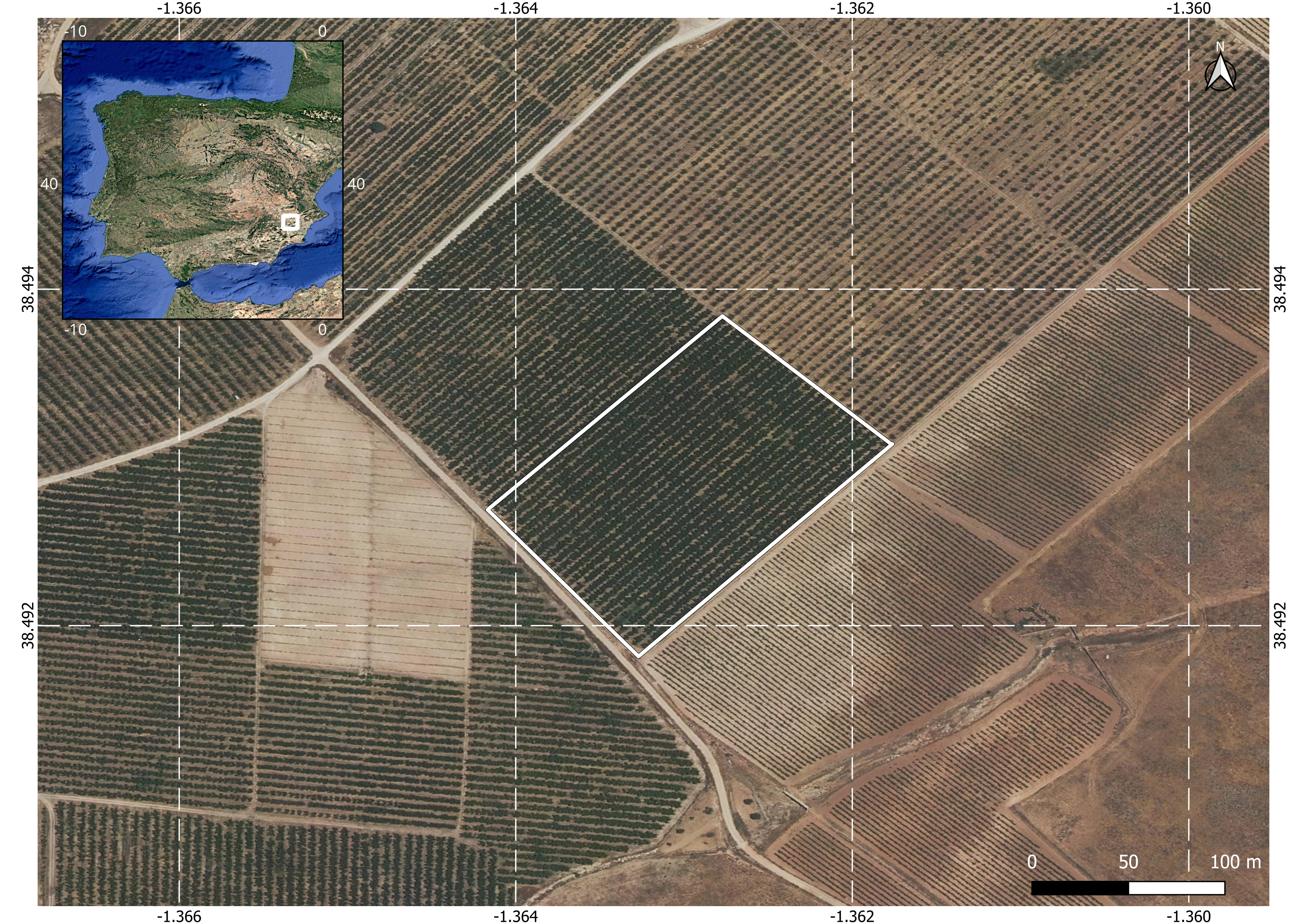}
\caption{Georeferenced position of the study plot designated for data acquisition and experimental analysis.\label{fig:FPlocation}}
\end{figure}

The local dataset was collected at a single field site: a 2~ha commercial peach orchard (Figure~\ref{fig:FPlocation}) located in Jumilla, in the Murcia region of Spain. The region has a semi-arid Mediterranean climate, characterized by dry, hot summers and rainy cold winters \citep{medda2022influence}, conditions that strongly influence the expression of abiotic and biotic stress in the crop. The orchard was established in 2010 with Grocivac~2 (IVIA) cultivar trees grafted onto GF677 rootstock, planted at a spacing of 5.5~m $\times$ 3~m.

\begin{figure}[H]
\centering
\subfloat[Original image taken from field]{\includegraphics[width=7cm,height=5cm,keepaspectratio]{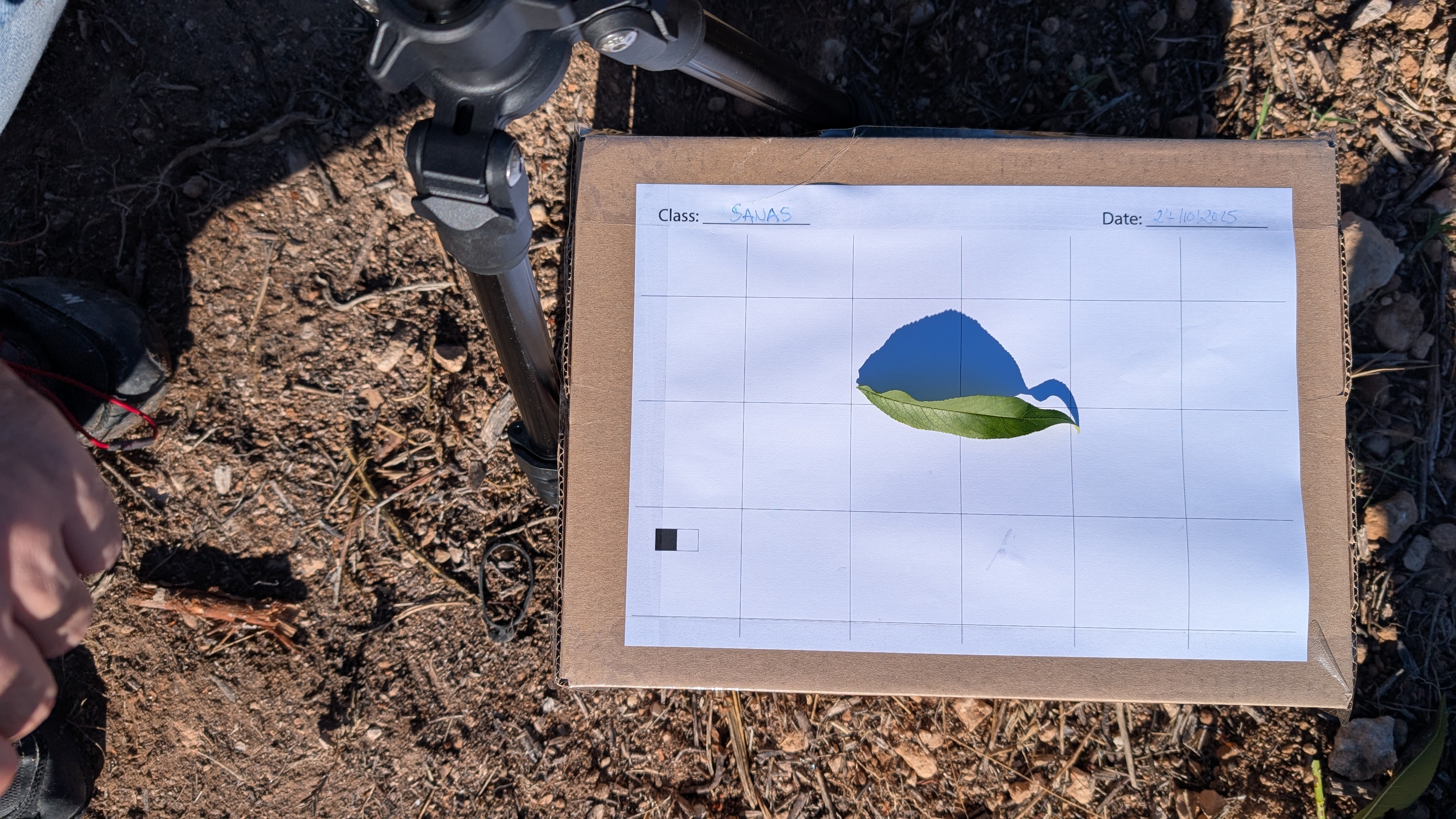}}
\hfill
\subfloat[Same leaf segmented and cropped for our dataset]{\includegraphics[width=7cm,height=5cm,keepaspectratio]{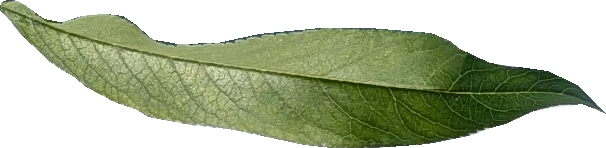}}
\caption{Representative examples of local leaf images. In this case a healthy leaf and how we cropped and processed it: (\textbf{a}) original image taken from field; (\textbf{b}) same leaf segmented and cropped for our dataset.\label{fig:two_leaves_example}}
\end{figure}

The first portion of the dataset consists of 26 in-situ images acquired in 2024 as part of a previous project. These images were manually segmented to isolate individual leaves, resulting in 73 healthy leaf samples. Given that this project was not initially focused on peach leaf pathology, the available imagery was limited. To mitigate this, a second image-collection campaign was carried out during the 2025 growing season at the same site using a camera module equipped with a 1/1.3'' sensor (50~MP, 1.2~$\upmu$m pixel pitch) and a fixed 24~mm equivalent focal length lens with f/1.7 aperture. This campaign yielded 107 additional images, including samples of healthy leaves, abiotic stress, chewing insects and mechanical damage. The number of collected images was constrained by both the limited occurrence of non-healthy conditions during the acquisition window and the fact that the campaign was conducted under late-season field conditions, where the orchard was already under established phytosanitary and biological control management. Consequently, certain stressors, specifically \textit{bacterial spot} and \textit{mite presence}, were not observed during the sampling period. Rather than artificially introducing external samples or simulated instances of these classes, the dataset was restricted to naturally occurring conditions to preserve ecological validity and avoid distributional inconsistencies. Nevertheless, this collection substantially increased the physiological diversity of the dataset, which was the primary objective of the second campaign. The final class distribution for the combined dataset is provided in Table~\ref{tab:class_distribution_local}, and an example of the acquisition and processing workflow is shown in Figure~\ref{fig:two_leaves_example}.

\begin{table}[H]
\centering
\caption{Class distribution of annotated leaf images from local field.\label{tab:class_distribution_local}}
\begin{tabularx}{0.6\textwidth}{lC}
\toprule
\textbf{Class} & \textbf{Number of Images}\\
\midrule
bacterial\_spot & 0\\
mechanical\_stress & 44\\
abiotic\_stress & 22\\
mite\_presence & 0\\
healthy & 107\\
chewing\_insect & 7\\
\midrule
\textbf{Total} & \textbf{180}\\
\bottomrule
\end{tabularx}
\end{table}

As we have mentioned, only four of the six classes present in public datasets were identified in our local data. Accordingly, we divided our experiments into two stages. First, we established a benchmark using the complete public dataset. Subsequently, we performed supervised domain adaptation by fine-tuning selected models on the four classes available in the local domain. This approach enables targeted adaptation without modifying the model architectures, ensuring a fair comparison and reproducibility. Our objective is to improve model performance on local data while preserving detection capabilities on the original (public) domain.

Additionally, a class imbalance is observed in the public dataset, where categories such as mechanical stress, mite presence, and chewing insects are underrepresented compared to the dominant control class (healthy leaves). A similar imbalance is present in the local dataset. To address this issue, we promote improved detection of minority classes and penalize models that underperform in this regard. The strategies adopted to mitigate this imbalance are described in Sections~\ref{dataprocessing}, \ref{Data_Augmentation} and~\ref{Experimental_Setup}.

\subsection{Data Preprocessing}\label{dataprocessing}

To mitigate the impact of class imbalance, an initial stratified 80/20 training-test split was employed for the public dataset (Table~\ref{tab:class_distribution}). To rigorously evaluate model generalization, the 80\% training partition was subjected to a repeated stratified cross-validation strategy, consisting of 5 folds repeated 3 times. This configuration, complemented by data augmentation and the application of dynamically computed balanced class weights during training, ensures that the models receive sufficient and equitable supervision from minority class samples. This is critical for learning the discriminative spatial and channel features required for accurate classification. The local dataset (Table~\ref{tab:class_distribution_local}) was initially utilized as an independent test set to quantify the domain gap of models trained exclusively on public data. To this end, a 80/20 split configuration was performed, and only the test partition was used in the first phase of evaluation. Subsequently, the remaining 80\% of the data was repurposed as a training set using the same repeated stratified cross-validation. This was meant to enable cross-domain transfer learning, with the objective of mitigating distributional shift and improving model performance in the local deployment context. This experimental design intentionally simulates a low-data regime, where the model must adapt to the specific morphological variations of the local field with a limited pool of samples. By focusing on this challenging scenario, we evaluate the robustness of the proposed models in generalizing from broad public databases to specific, small-scale agricultural environments where classes are critically underrepresented.

These images were preprocessed in the following ways. First, they were cropped and resized to multiple standardized dimensions---specifically 192 $\times$ 192 and 224 $\times$ 224 pixels---to systematically evaluate the impact of input resolution on model performance while establishing a fair common baseline. Since several experimental configurations were explored for each architecture, including different input image sizes, unless explicitly stated otherwise, the Results section reports only the performance obtained with the best-performing image size configuration over the validation folds for each model. After this, these images were preprocessed following each model's features. For that purpose, the function \textit{preprocess\_input} from Keras \citep{keras} was used for each Keras model. In order to use the pre-trained models efficiently, these images were standardized using the mean and standard deviation of ImageNet \citep{ImageNet}, the dataset originally used for training the aforementioned algorithms.

\subsection{Data Augmentation}\label{Data_Augmentation}

To improve the model's robustness and reduce overfitting, a custom data augmentation pipeline was applied during training, aimed at increasing variability in the input images while preserving class semantics. The following techniques were employed:

\begin{itemize}
\item Random rotations of up to $\pm 10^\circ$ with a probability of $p = 0.5$;
\item Random resized crops, with scaling between 90\% and 100\%, and aspect ratio variations of up to 10\% with a probability of $p = 0.3$;
\item Random horizontal flips with a probability of $0.5$ and vertical flips with a probability of $0.2$;
\item Random brightness and contrast adjustments of up to $\pm 8$\% with $p = 0.2$;
\item Random hue, saturation, and value shifts with small limits (hue $\pm 3$, saturation $\pm 5$, value $\pm 4$, $p = 0.2$);
\item Small random shifts and scaling (up to 5\%) without rotation ($p = 0.3$);
\item Resizing to the target input dimensions.
\end{itemize}

This augmentation strategy was applied on-the-fly during training using the Albumentations library \citep{Albumentations} in CNN models.

With this, we raised the number of images in the training subset to obtain at least 400 images for each class, enhancing the dataset for a more balanced distribution.

\subsection{Deep Learning Models}

The models considered in this study were selected to cover a range of state-of-the-art CNN-based architectures, spanning different families to evaluate the trade-offs between accuracy and computational cost. Additionally, we propose hybrid models that integrate CBAM attention into light, medium and large CNN backbones, aiming to explore more efficient architectures for leaf classification.

\subsubsection{MobileNet Family}

The MobileNet family was originally proposed for mobile vision applications \citep{MobileNetpaper}. MobileNet uses depthwise separable convolutions, factorizing standard convolutions into a depthwise convolution followed by a point-wise 1 $\times$ 1 convolution, significantly reducing computational cost and model size. MobileNetV2 \citep{MobileNetv2} extends this design with residual bottleneck layers and linear bottlenecks, improving feature representation while maintaining efficiency. MobileNetV3 \citep{MobileNetv3} further introduces Squeeze-and-Excitation (SE) blocks \citep{SEpaper} and the hard-swish activation, combining automated neural architecture search with manual optimization to enhance feature representation in a lightweight model. We selected MobileNetV2 and MobileNetV3-Large for our experiments to compare the previously described modifications and to evaluate their performance against our proposed hybrid architectures.

\subsubsection{EfficientNet Family}

EfficientNet \citep{EfficientNet} was designed to explore model scaling and network balancing. EfficientNet-B0 serves as the baseline, built with mobile inverted bottleneck blocks and SE optimization, similar to MobileNet. Using the compound scaling method, larger variants (B1--B7) are generated by jointly scaling depth, width, and resolution. For our experiments, we included EfficientNet-B0, B3, and B5 to assess the impact of model size on leaf classification performance.

\subsubsection{ResNet Family}

ResNet50 and ResNet101 \citep{ResNet} introduce residual connections to mitigate vanishing gradient issues in deeper networks, which are crucial for working with challenging datasets like ours. ResNet50 and ResNet101 provide progressively deeper models for evaluating the effect of increased depth and representational capacity on classification performance.

\subsubsection{VGG Family}

The VGG family \citep{VGG} consists of deep CNNs with uniform 3 $\times$ 3 convolutional filters stacked in increasing depths, followed by fully connected layers at the end. VGG16 and VGG19 have 16 and 19 layers, respectively, with a very high number of parameters. These models represent more traditional deep CNN architectures, emphasizing depth and simplicity of design. Despite their large size, they have been widely used as strong baselines in image classification tasks due to their straightforward structure and proven performance. We included them in order to compare more traditional architectures with novel approaches.

\subsubsection{InceptionV3}

InceptionV3 \citep{InceptionV3} is based on the Inception architecture, which combines multiple convolutional kernel sizes in parallel within a single module to capture features at different scales. It includes auxiliary classifiers for regularization and factorized convolutions to reduce computational costs.

\subsubsection{DenseNet Family}

The DenseNet family \citep{DenseNet} introduces dense connectivity between layers, where each layer receives as input the feature maps of all preceding layers and passes its own feature maps to all subsequent layers. This design encourages feature reuse, improves gradient flow, and mitigates the vanishing gradient problem, allowing very deep networks to be trained efficiently. DenseNet architectures are typically denoted as DenseNet-\{number of layers\}, with DenseNet121, DenseNet169, and DenseNet201 being common variants. For our study, we selected DenseNet121 providing a compact yet powerful network that balances depth and computational cost.

\subsection{Proposed Models}

In this study, we propose implementing the Convolutional Block Attention Module (CBAM) to enhance accuracy without incurring critical computational costs.

\subsubsection{Convolutional Block Attention Module (CBAM)}

CBAM was introduced in \citep{CBAMpaper} as a lightweight and generic attention mechanism to refine convolutional feature maps. It extends the Squeeze-and-Excitation (SE) module, which applies channel-wise attention, by additionally incorporating spatial attention and using both average and maximum pooling operations. In all experiments, we set the reduction ratio $r = 16$.

Given an input feature map $X \in \mathbb{R}^{C \times H \times W}$, CBAM sequentially infers two types of attention: channel attention and spatial attention.

\textbf{Channel Attention.} Two descriptors are obtained through global average and max pooling:

\begin{equation}
f^{avg}_c = \frac{1}{H W}\sum_{i=1}^H \sum_{j=1}^W X_{c,i,j}, \quad
f^{max}_c = \max_{i,j} X_{c,i,j}
\end{equation}

Both are passed through a shared MLP with \textit{reduction ratio} $r$ in order to maintain an efficient architecture and control model size:

\begin{equation}
M_c(f) = W_1(\text{ReLU}(W_0 f)),
\end{equation}

where $W_0 \in \mathbb{R}^{\tfrac{C}{r} \times C}$, $W_1 \in \mathbb{R}^{C \times \tfrac{C}{r}}$. The channel attention map is then:

\begin{equation}
A_c = \sigma\big( M_c(f^{avg}_c) + M_c(f^{max}_c) \big),
\end{equation}

with $A_c \in \mathbb{R}^{C \times 1 \times 1}$. The refined feature map is:

\begin{equation}
X' = A_c \odot X
\end{equation}

\textbf{Spatial Attention.} From $X'$, descriptors are obtained by channel-wise pooling:

\begin{equation}
f^{avg}_s(i,j) = \frac{1}{C}\sum_{c=1}^C X'_{c,i,j}, \quad
f^{max}_s(i,j) = \max_c X'_{c,i,j}
\end{equation}

These maps are concatenated and passed through a $7 \times 7$ convolution:

\begin{equation}
M_s = \text{Conv}_{7 \times 7}\big([f^{avg}_s; f^{max}_s]\big),
\end{equation}

yielding the spatial attention:

\begin{equation}
A_s = \sigma(M_s), \quad A_s \in \mathbb{R}^{1 \times H \times W}
\end{equation}

Finally, the output is:

\begin{equation}
Y = A_s \odot X'
\end{equation}

\subsubsection{Implementation Details}

In the original CBAM paper, the module is placed after each convolutional block. In our implementation, CBAM is attached to the output of the last convolutional stage of each backbone, before global average pooling and the classification head. This design choice enriches the final representation while preserving the lightweight nature of the module. After the sequential process of introducing channel and spatial attention, we have generated an output $Y$, which must then be transformed into a class prediction. For this purpose, we apply 2D global average pooling (GAP) to the CBAM-refined output $Y$, reducing it to a compact feature vector. This is followed by two dense layers with 256 neurons and ReLU activation, with a dropout layer ($p = 0.5$) applied between them to mitigate overfitting, and a final softmax layer for classification. This architecture emphasizes efficiency while maintaining representational capacity, producing lightweight models with a reduced risk of overfitting.

\begin{figure}[H]
\centering
\includegraphics[width=\textwidth]{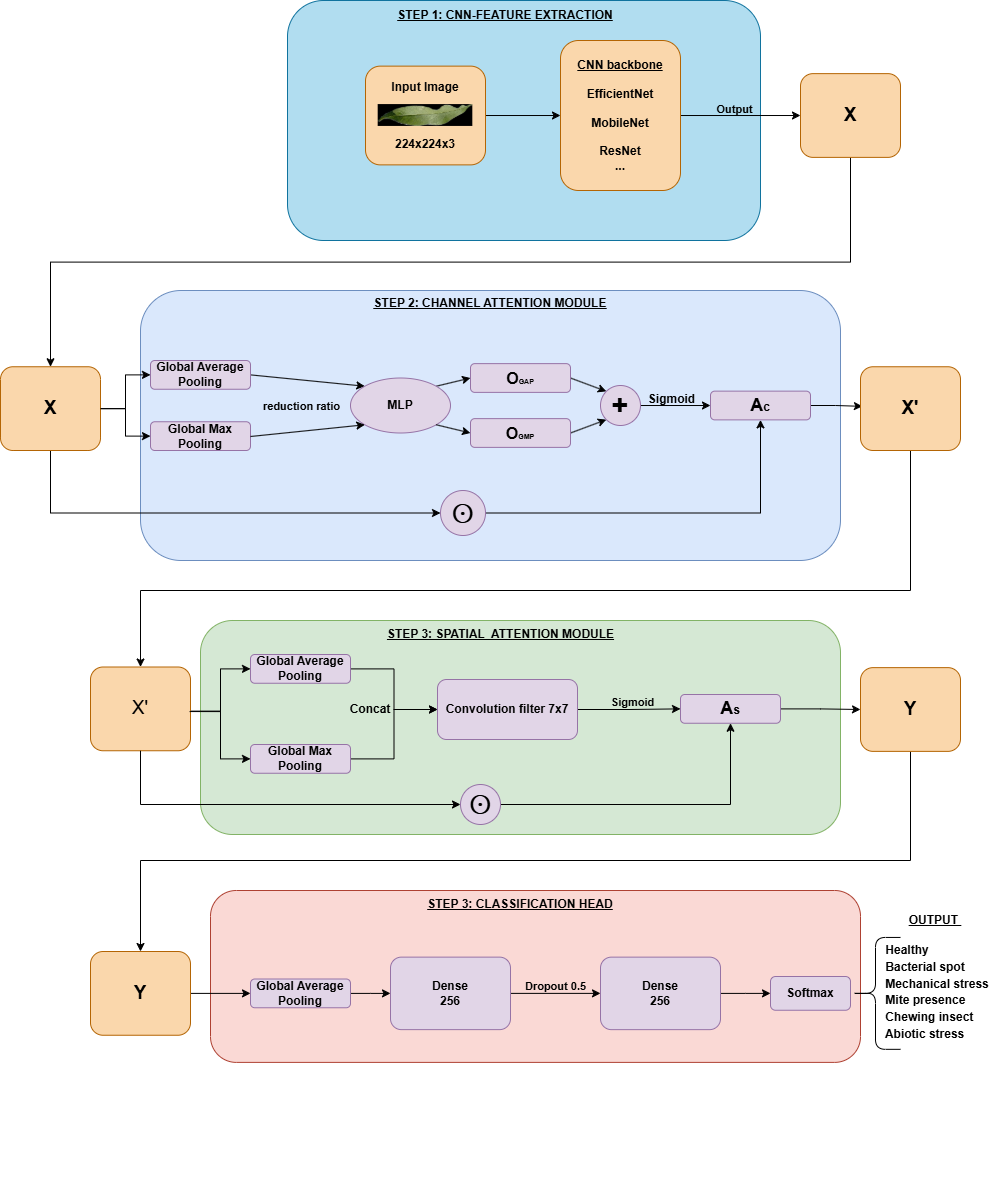}
\caption{CBAM implementation scheme. Parameters H, W and C depend on each backbone output.\label{fig:CBAM_scheme}}
\end{figure}

This configuration allows us to compare enhanced CBAM models with base models, providing insights into the trade-off between efficiency and feature richness when integrating CBAM into CNN backbones. The described header, together with the CBAM block, was integrated into the backbones of the aforementioned deep learning models. A summarized diagram of these architectures is shown in Figure~\ref{fig:CBAM_scheme}.

In order to understand the impact on the model size, see Table~\ref{tab:model_sizes} where we included some examples of parameter increase for models of different sizes. The relative increase in model size is practically negligible with respect to the original number of parameters, as we aim to maintain computational efficiency in the models.

\begin{table}[H]
\centering
\caption{Model sizes (number of parameters) for different classification heads.\label{tab:model_sizes}}
\begin{tabularx}{0.8\textwidth}{>{\raggedright\arraybackslash}XCC}
\toprule
\textbf{Model} & \textbf{Original} & \textbf{Original + CBAM}\\
\midrule
DenseNet121 & 12,765,780 & 13,182,072\\
EfficientNetB0 & 10,181,871 & 11,002,089\\
EfficientNetB3 & 27,494,847 & 28,673,220\\
EfficientNetB5 & 73,837,435 & 75,885,901\\
InceptionV3 & 48,774,964 & 50,171,812\\
MobileNetV2 & 5,839,316 & 6,611,708\\
MobileNetV3Large & 7,949,876 & 8,497,394\\
ResNet50 & 57,994,132 & 60,016,672\\
ResNet101 & 88,221,076 & 90,243,616\\
VGG16 & 33,602,388 & 33,821,280\\
VGG19 & 43,631,700 & 43,850,592\\
\bottomrule
\end{tabularx}
\end{table}

\subsection{Experiment Setup}\label{Experimental_Setup}

The models were trained under a unified setup to ensure a fair comparison across different architectures. We employed the Adam optimizer \citep{Adam} with an initial learning rate of $1 \times 10^{-3}$. The training process was divided into two stages. In the first stage, the models were trained for 100 epochs with the backbone frozen, incorporating class weights to address class imbalance and using ImageNet pre-trained weights. Early stopping and learning rate reduction on plateau were applied to mitigate overfitting. In the second stage, training continued for an additional 120 epochs, with the backbone partially unfrozen according to an unfreeze ratio of 0.3 and a starting learning rate of $1 \times 10^{-5}$. The same overfitting mitigation strategies were maintained.

The loss function used in both stages was the class-weighted categorical cross-entropy loss ($L$), designed to mitigate the impact of class imbalance by assigning higher importance to minority classes. This function evaluates the discrepancy between the predicted probability distribution and the true class labels as follows:

\begin{equation}
L = - \frac{1}{N} \sum_{i=1}^{N} \sum_{c=1}^{C} w_c \, y_{i,c} \log(\hat{y}_{i,c})
\end{equation}

where $N$ is the number of samples, $C$ is the number of classes, $y_{i,c}$ is a binary indicator (0 or 1) denoting whether sample $i$ belongs to class $c$, $\hat{y}_{i,c}$ is the predicted probability for class $c$ after the softmax activation, and $w_c$ is the class-specific weight inversely proportional to class frequency. This weighting scheme penalizes the misclassification of underrepresented classes more strongly, encouraging the model to learn more balanced decision boundaries. Combined with oversampling through data augmentation, this approach helps further alleviate the effects of class imbalance during training.

Input images were resized to $192 \times 192 \times 3$ and $224 \times 224 \times 3$ to evaluate the effect of resolution on model performance. Preprocessing was applied according to the requirements of each backbone, using the corresponding model-specific preprocessing functions. Table~\ref{tab:setup_training} summarizes the complete training configuration. This setup was carefully selected based on preliminary experiments aimed at identifying optimal hyperparameters for our specific dataset, and was consistently applied across all cross-validation iterations.

\begin{table}[H]
\centering
\caption{Set up for training models.\label{tab:setup_training}}
\begin{tabularx}{\textwidth}{lXX}
\toprule
\textbf{Parameter} & \textbf{Stage 1} & \textbf{Stage 2}\\
\midrule
Optimizer & Adam & Adam\\
Learning rate & 0.001 & 0.00001\\
Backbone & Frozen & Unfreeze ratio of 0.3\\
Epochs & 100 & 120\\
Loss Function & Weighted Categorical Cross-Entropy & Weighted Categorical Cross-Entropy\\
Early Stopping & Patience = 3 & Patience = 4\\
ReduceLROnPlateau & Factor = 0.5, Patience = 2 & Factor = 0.5, Patience = 2\\
Input size (pixels) & \multicolumn{2}{c}{224 $\times$ 224 $\times$ 3, 192 $\times$ 192 $\times$ 3}\\
Pretrained weights & \multicolumn{2}{c}{ImageNet}\\
Augmentation & \multicolumn{2}{c}{Described in Section~\ref{Data_Augmentation}}\\
\bottomrule
\end{tabularx}
\end{table}

The evaluation metrics considered in this study are Accuracy, F1-score, macro Precision, macro Recall, and macro F1-score. Here, the macro variants refer to the standard metrics averaged across classes, giving equal weight to each class regardless of its representation in the dataset. This approach allows us to account for class imbalance and place greater emphasis on minority classes.

Accuracy measures the overall proportion of correctly classified samples. Precision and recall are computed as follows:

\begin{equation}
\text{Precision} = \frac{TP}{TP + FP}
\end{equation}

\begin{equation}
\text{Recall} = \frac{TP}{TP + FN}
\end{equation}

where $TP$, $FP$, and $FN$ denote true positives, false positives, and false negatives, respectively. The F1-score is the harmonic mean of precision and recall, defined as:

\begin{equation}
\text{F}_1 = 2 \times \frac{\text{Precision} \times \text{Recall}}{\text{Precision} + \text{Recall}}
\end{equation}

Finally, the macro versions of the aforementioned metrics are obtained by averaging the corresponding class-wise values across all classes, as defined below:

\begin{align}
\text{Macro F1} &= \frac{1}{C}\sum_{i=1}^{C} (F_1)_i \\
\text{Macro Precision} &= \frac{1}{C}\sum_{i=1}^{C} \text{Precision}_i \\
\text{Macro Recall} &= \frac{1}{C}\sum_{i=1}^{C} \text{Recall}_i
\end{align}

These metrics collectively provide a comprehensive evaluation of model performance, particularly under imbalanced class distributions. All experiments were implemented in Python and executed on a dedicated HPC server equipped with an NVIDIA H100 NVL GPU (94~GB HBM3), 40 CPU cores, and 32~GB of RAM. Model training, evaluation, and data preprocessing were performed using standard deep learning libraries, including TensorFlow \citep{Tensorflow} and PyTorch \citep{Pytorch}.

\section{Results}

The experiments conducted can be divided into two parts. First, we establish a benchmark assessing which models are the most suitable for the developed task, depending on their computational cost and performance metrics, and evaluate how the implementation of CBAM in different models affects their classification capability. Secondly, we report on Transfer Learning, showing how different fine-tuning techniques affect performance on the local dataset.

\subsection{Performance Benchmark and CBAM Enhancement}

\subsubsection{Benchmark}

In the first conducted experiment, the setup described in Table~\ref{tab:setup_training} was used to train the models, aiming to establish a robust benchmark.

The results are summarized in Table~\ref{tab:backbones_summary_updated}, where the best-performing model within each category is highlighted in bold. The reported configurations correspond to the optimal settings obtained after the full cross-validation process, considering both weighted (overall performance) and macro-averaged (class-balanced performance) metrics.

\begin{table}[H]
\centering
\small
\caption{Summary of the best configuration per model. Weighted metrics (W) represent overall performance, while macro metrics (M) capture class-balanced behavior. Models are grouped according to number of parameters.\label{tab:backbones_summary_updated}}
\begin{tabularx}{\textwidth}{lCCCCCC}
\toprule
\textbf{Model} & \textbf{Img Size} & \textbf{F1 (W)} & \textbf{Accuracy (W)} & \textbf{Precision (M)} & \textbf{Recall (M)} & \textbf{F1 (M)}\\
\midrule
\multicolumn{7}{c}{\textit{Light Models (<15M params)}}\\
\midrule
MobileNetV2 & 192 & 0.899 & 0.898 & 0.778 & 0.730 & 0.733\\
MobileNetV3Large & 224 & 0.899 & 0.901 & 0.789 & 0.743 & 0.764\\
EfficientNetB0 & 224 & 0.927 & 0.929 & 0.861 & 0.786 & 0.817\\
DenseNet121 & 192 & 0.923 & 0.926 & 0.840 & 0.787 & 0.801\\
\midrule
\multicolumn{7}{c}{\textit{Medium Models (15M--60M params)}}\\
\midrule
EfficientNetB3 & 192 & 0.921 & 0.915 & 0.828 & 0.819 & 0.806\\
ResNet50 & 192 & 0.913 & 0.908 & 0.822 & 0.781 & 0.764\\
InceptionV3 & 224 & 0.889 & 0.894 & 0.869 & 0.701 & 0.762\\
VGG16 & 224 & 0.905 & 0.908 & 0.796 & 0.722 & 0.750\\
VGG19 & 192 & 0.901 & 0.898 & 0.817 & 0.786 & 0.789\\
\midrule
\multicolumn{7}{c}{\textit{Large Models (>60M params)}}\\
\midrule
EfficientNetB5 & 192 & 0.922 & 0.919 & 0.876 & 0.832 & 0.837\\
ResNet101 & 192 & 0.919 & 0.919 & 0.795 & 0.779 & 0.778\\
\midrule
\multicolumn{7}{c}{\textit{CBAM/Attention-Enhanced Models}}\\
\midrule
EfficientNetB3 + CBAM & 192 & 0.915 & 0.915 & 0.828 & 0.814 & 0.819\\
InceptionV3 + CBAM & 224 & 0.909 & 0.912 & 0.823 & 0.776 & 0.791\\
EfficientNetB5 + CBAM & 224 & 0.936 & 0.933 & 0.838 & 0.871 & 0.849\\
\bottomrule
\end{tabularx}
\end{table}

Lightweight models generally provide strong baseline performance despite their reduced computational complexity. Among them, EfficientNetB0 achieves the best overall results, reaching an accuracy of 92.9\% and a weighted F1-score of 92.7\%, closely followed by MobileNetV3Large and MobileNetV2. However, differences between weighted and macro-averaged metrics reveal a consistent performance gap across classes, suggesting that certain minority categories are not being equally well represented by these architectures. In particular, macro F1 scores remain notably lower than weighted ones, indicating the presence of class imbalance effects. These observations motivate the use of attention mechanisms such as CBAM to improve feature discrimination, particularly for underrepresented classes.

Among medium-sized architectures, DenseNet121 achieves the best overall performance in terms of weighted metrics, reaching 92.6\% accuracy and a weighted F1-score of 92.3\% with 12.8M parameters, confirming its efficiency in feature reuse and gradient propagation. EfficientNetB3 also demonstrates competitive performance, particularly in macro recall ($0.819$) and macro F1 ($0.806$), indicating a better balance across classes. In contrast, architectures such as ResNet50 and InceptionV3 do not consistently outperform simpler models, reinforcing the idea that increased depth does not necessarily translate into better generalization when training data is limited.

For large-scale models, EfficientNetB5 achieves the best overall performance, with a weighted F1-score of 92.2\% and strong macro-level results (Precision: $0.876$, Recall: $0.832$, F1: $0.837$), making it the most robust architecture among non-attention-based models. In contrast, deeper architectures such as ResNet101 (weighted F1: $0.919$) and VGG variants (weighted F1: $0.901$--$0.905$) do not provide substantial improvements and, in some cases, exhibit marginal performance degradation compared to EfficientNetB5, further supporting the hypothesis that model capacity must be carefully matched to dataset size and complexity to avoid overfitting.

\begin{figure}[H]
\centering
\subfloat[EfficientNetB0]{\includegraphics[width=7cm]{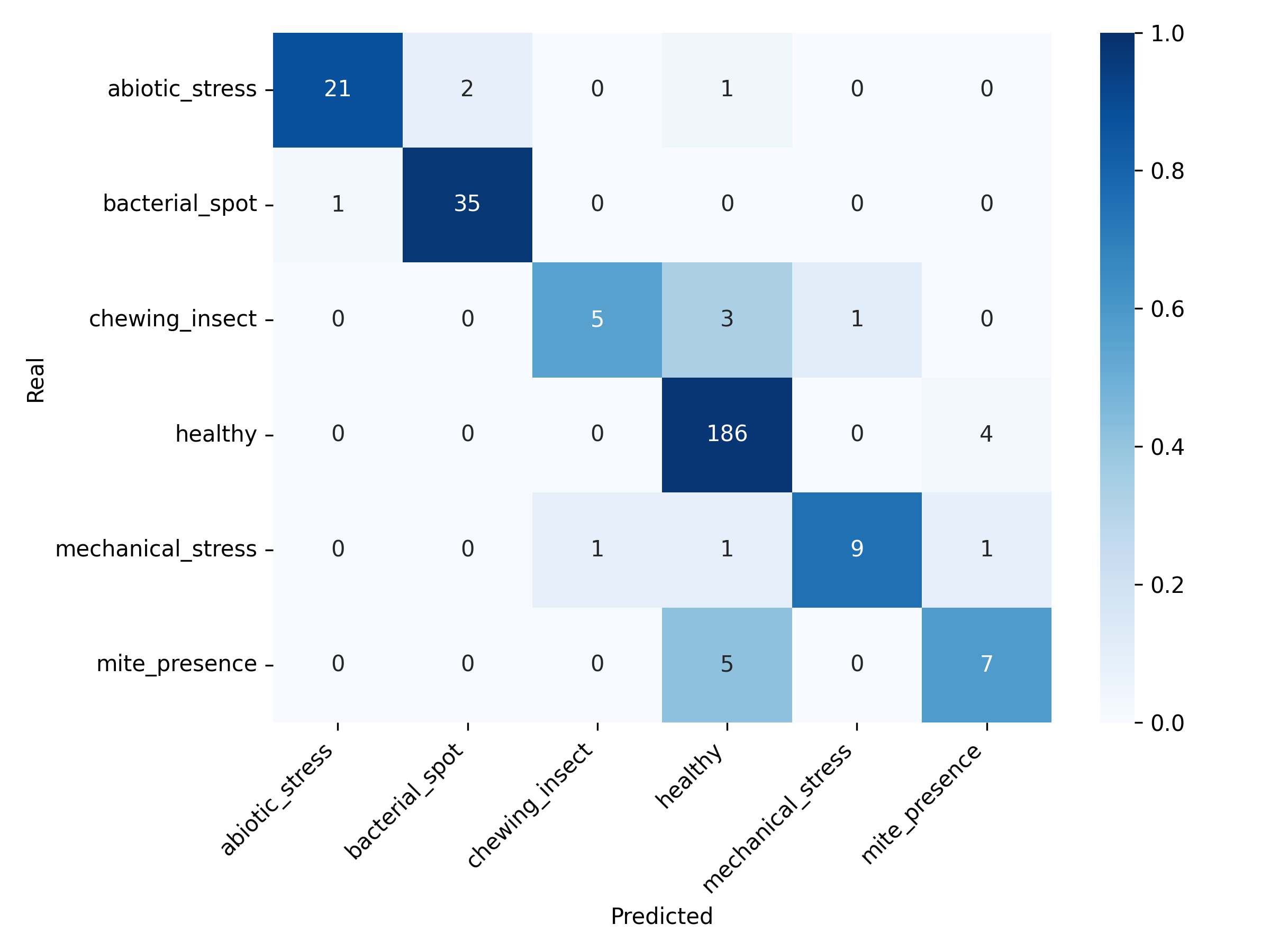}}
\hfill
\subfloat[DenseNet121]{\includegraphics[width=7cm]{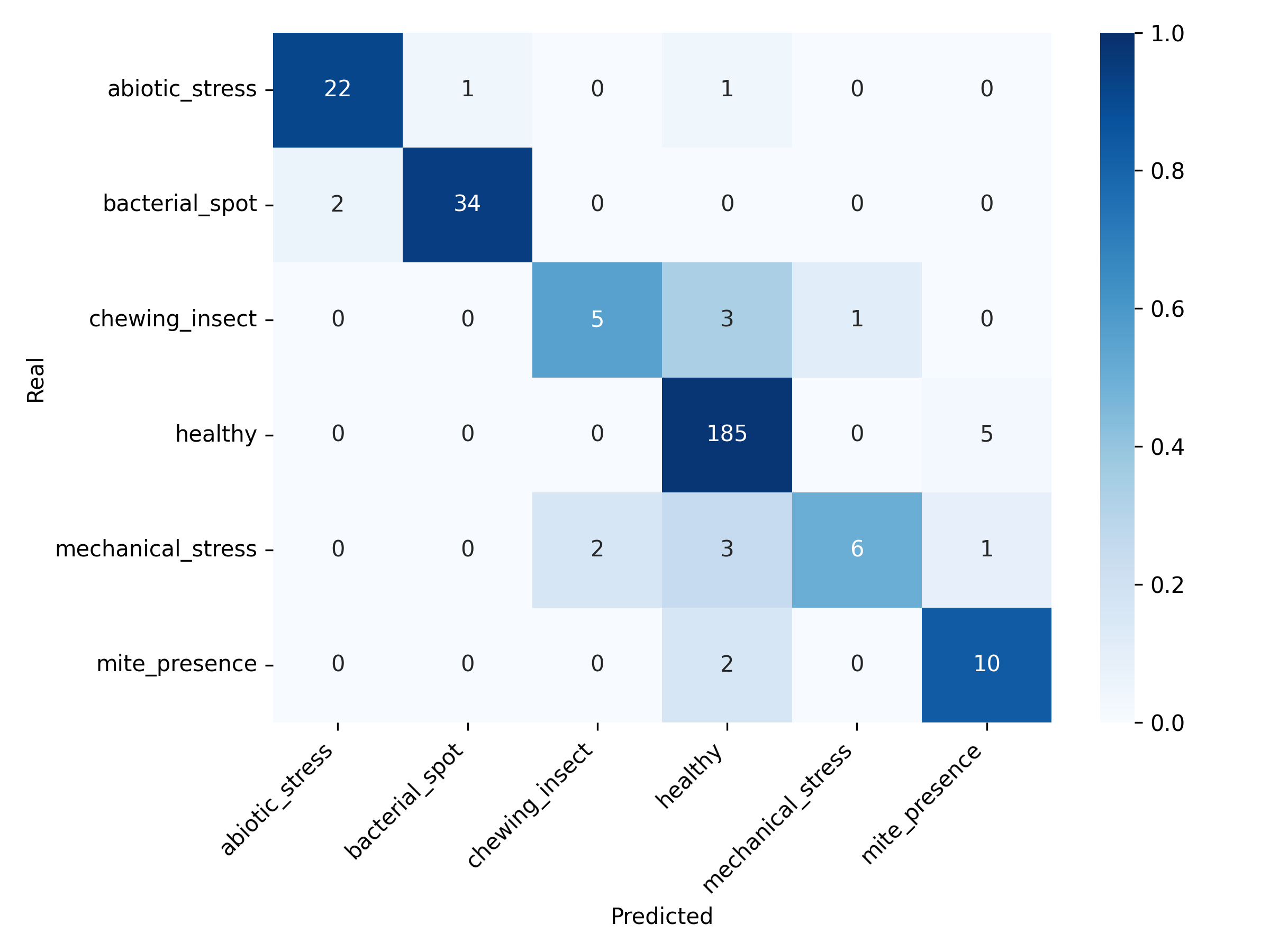}}\\
\subfloat[EfficientNetB5]{\includegraphics[width=7cm]{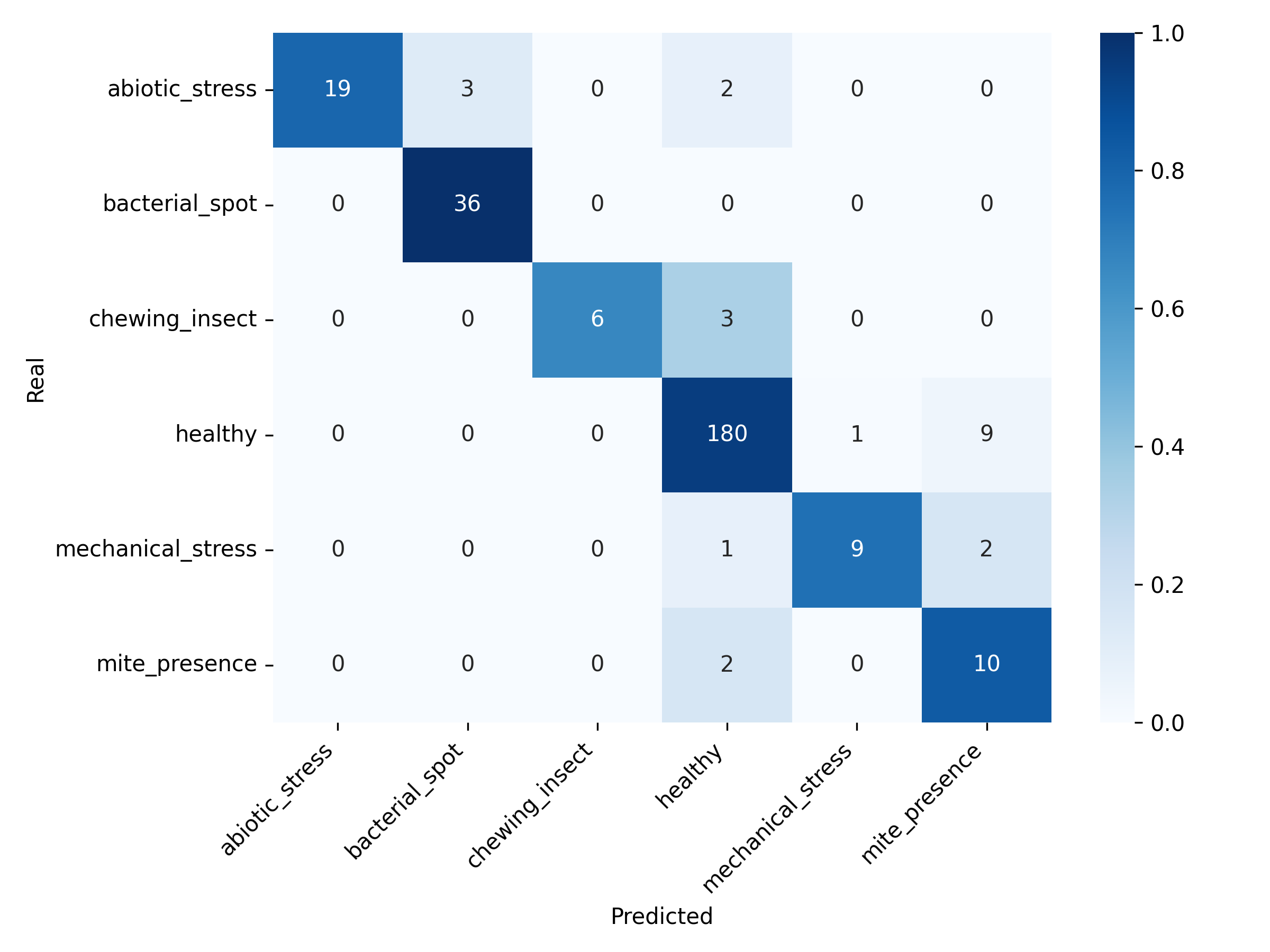}}
\hfill
\subfloat[EfficientNetB5 + CBAM]{\includegraphics[width=7cm]{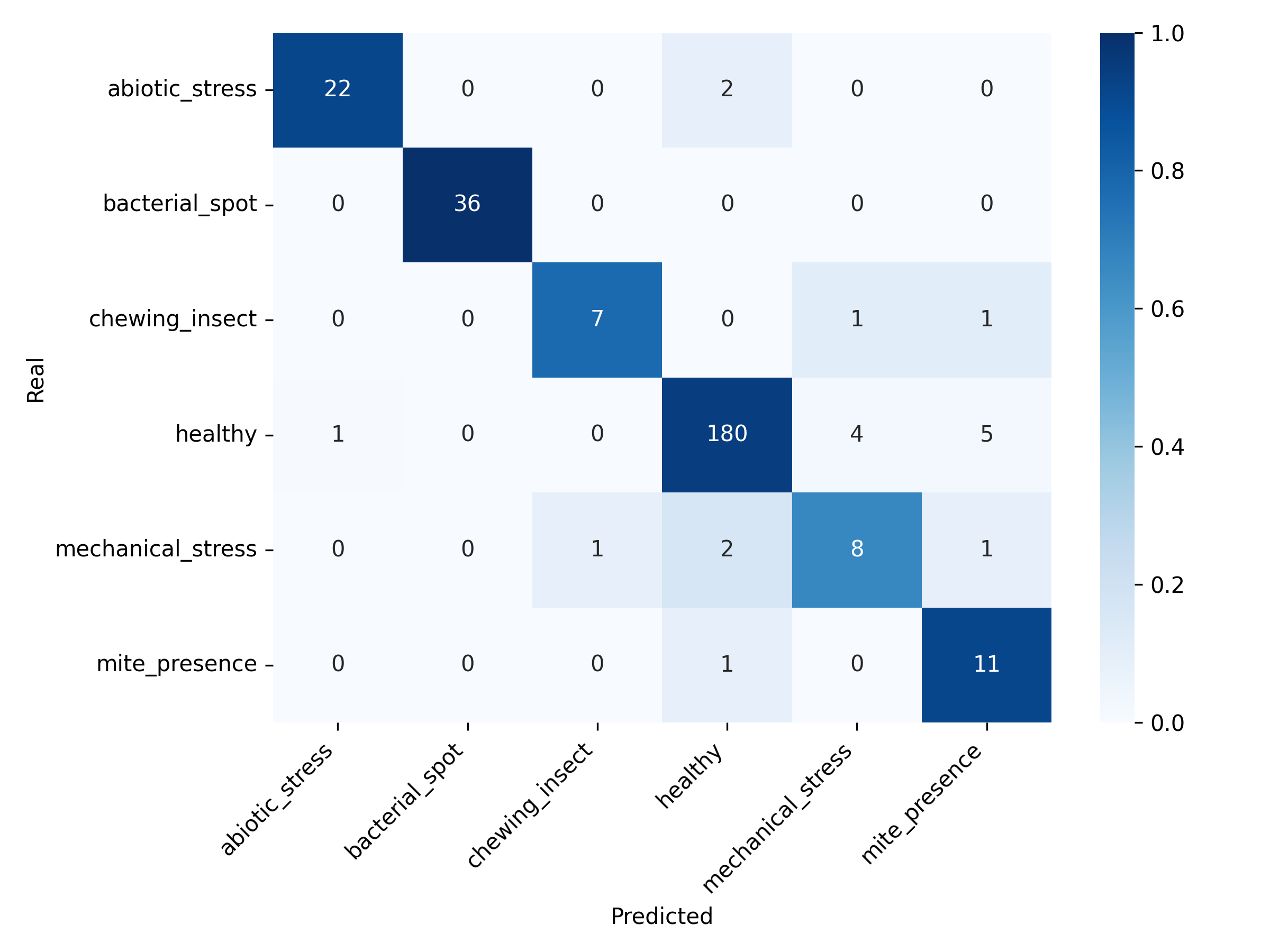}}
\caption{Confusion matrices of representative models: (\textbf{a}) EfficientNetB0 (light); (\textbf{b}) DenseNet121 (medium); (\textbf{c}) EfficientNetB5 (large); (\textbf{d}) EfficientNetB5 + CBAM (CBAM model). The cells show the number of examples classified in each category, while the color intensity is row-normalized (per class) to highlight relative correct and incorrect predictions for each class.\label{fig:conf_matrices_4_models}}
\end{figure}

Figure~\ref{fig:conf_matrices_4_models} shows the confusion matrices of representative models from each size group. Across models, three error patterns recur: (1) confusion between \textit{Bacterial Spot} and \textit{Abiotic Stress}; (2) bias towards the \textit{Healthy} class (the majority class); and (3) confusion between \textit{Chewing Insect} and \textit{Mechanical Stress}. Following this benchmarking phase, the same procedure was repeated for models enhanced with CBAM.

\subsubsection{Model Improvement via CBAMs}

After establishing the baseline benchmark, we incorporated Convolutional Block Attention Modules (CBAM) into selected backbone architectures to enhance feature representation and improve minority class detection. The consistent gap between weighted and macro-averaged metrics observed across all model families suggested that class imbalance remained a limiting factor, motivating the use of attention mechanisms to improve feature discrimination for underrepresented classes. The experimental setup was kept identical to the baseline configuration to ensure a fair comparison.

\begin{figure}[H]
\centering
\subfloat[EfficientNetB3 vs.\ EfficientNetB3 + CBAM]{\includegraphics[width=0.85\textwidth]{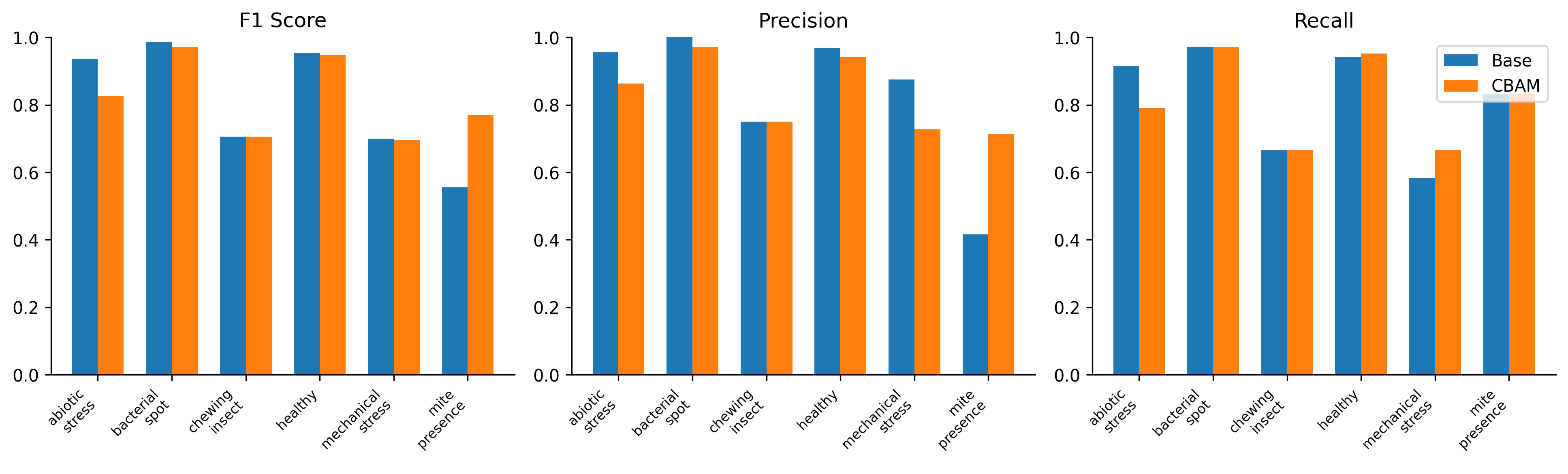}}\\
\subfloat[InceptionV3 vs.\ InceptionV3 + CBAM]{\includegraphics[width=0.85\textwidth]{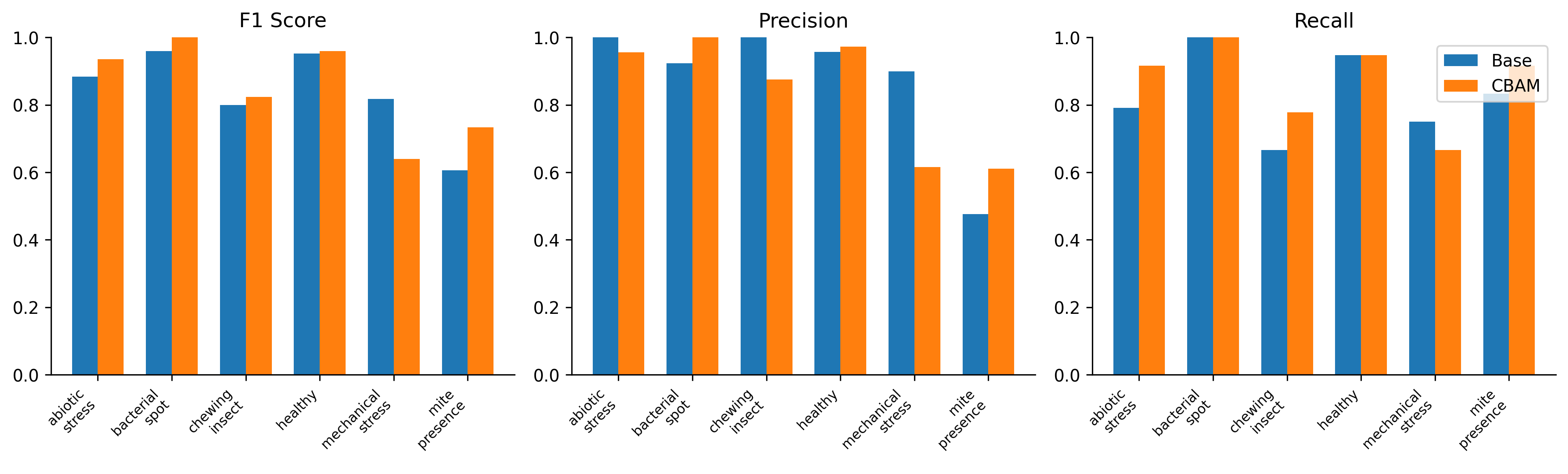}}\\
\subfloat[EfficientNetB5 vs.\ EfficientNetB5 + CBAM]{\includegraphics[width=0.85\textwidth]{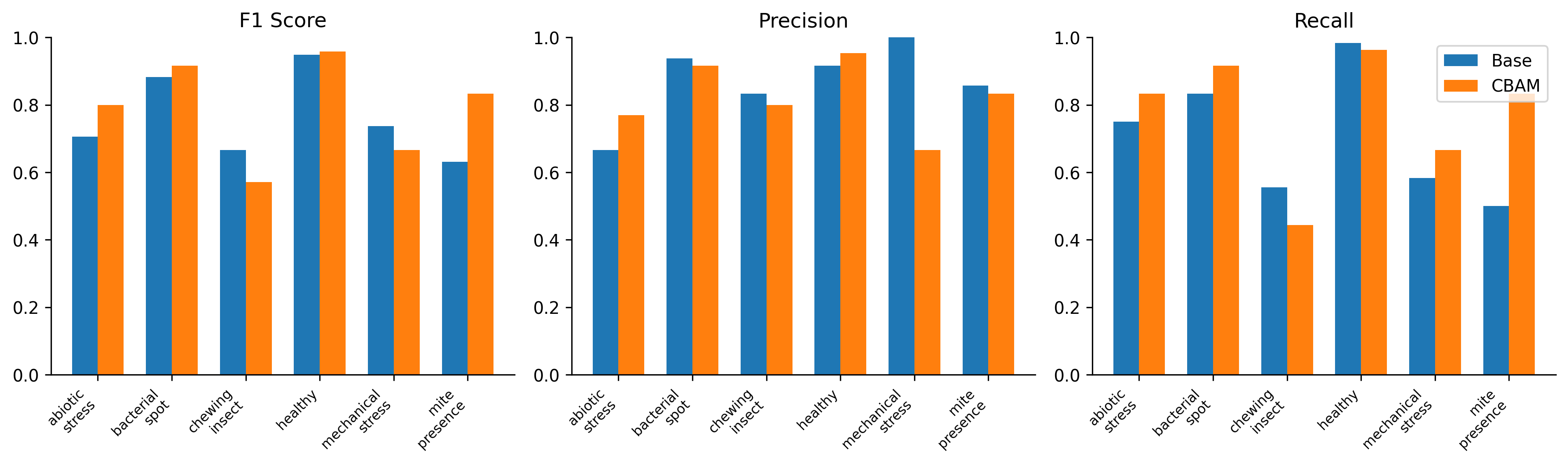}}
\caption{Class-wise comparison of F1, Recall, and Precision for EfficientNetB3, InceptionV3, and EfficientNetB5, contrasting their baseline and CBAM-enhanced versions. Each subplot shows one backbone model: (\textbf{a}) EfficientNetB3; (\textbf{b}) InceptionV3; (\textbf{c}) EfficientNetB5.\label{fig:class_metrics_comparison}}
\end{figure}

EfficientNetB3, InceptionV3, and EfficientNetB5 were selected for this analysis due to their strong baseline performance, allowing a comprehensive evaluation of CBAM's impact across model capacities. Table~\ref{tab:cbam_comparison} summarizes the aggregate performance differences, while Figure~\ref{fig:class_metrics_comparison} provides a class-wise breakdown of F1, Recall, and Precision for each model pair.

\begin{table}[H]
\centering
\caption{Performance comparison between base architectures and their counterparts integrated with CBAM for selected models.\label{tab:cbam_comparison}}
\begin{tabularx}{\textwidth}{lCCCCCC}
\toprule
\textbf{Metric} & \multicolumn{2}{c}{\textbf{EfficientNetB3}} & \multicolumn{2}{c}{\textbf{InceptionV3}} & \multicolumn{2}{c}{\textbf{EfficientNetB5}}\\
\cmidrule(lr){2-3} \cmidrule(lr){4-5} \cmidrule(lr){6-7}
& Base & CBAM & Base & CBAM & Base & CBAM\\
\midrule
F1 (W) & \textbf{0.921} & 0.915 & 0.889 & \textbf{0.909} & 0.922 & \textbf{0.936}\\
Accuracy (W) & 0.915 & \textbf{0.915} & 0.894 & \textbf{0.912} & 0.919 & \textbf{0.933}\\
Precision (M) & 0.828 & \textbf{0.828} & \textbf{0.869} & 0.823 & \textbf{0.876} & 0.838\\
Recall (M) & \textbf{0.819} & 0.814 & 0.701 & \textbf{0.776} & 0.832 & \textbf{0.871}\\
F1 (M) & 0.806 & \textbf{0.819} & 0.762 & \textbf{0.791} & 0.837 & \textbf{0.849}\\
\bottomrule
\end{tabularx}
\end{table}

The results in Table~\ref{tab:cbam_comparison} show that CBAM improves performance in most cases, though the nature and magnitude of the gains vary across architectures. EfficientNetB5 + CBAM achieves the strongest overall results, with weighted F1 increasing from $0.922$ to $0.936$ and macro F1 from $0.837$ to $0.849$. Notably, recall rises from $0.832$ to $0.871$, the largest absolute gain among the three models, suggesting that CBAM enables the backbone to better exploit its representational capacity for minority class detection. InceptionV3 + CBAM shows the most significant relative improvement: weighted F1 increases from $0.889$ to $0.909$ and macro F1 from $0.762$ to $0.791$, driven primarily by a substantial recall gain from $0.701$ to $0.776$. This indicates that CBAM is particularly effective for architectures that, without attention, exhibit a stronger bias towards majority classes. EfficientNetB3 + CBAM presents a more nuanced outcome: macro F1 improves from $0.806$ to $0.819$, but the weighted F1-score slightly decreases from $0.921$ to $0.915$.

The class-wise breakdown in Figure~\ref{fig:class_metrics_comparison} further clarifies these patterns, revealing that the impact of CBAM is heterogeneous across both architectures and classes.

For EfficientNetB3, the most notable improvement is observed in \textit{Mite Presence}, where performance increases substantially due to a strong gain in precision (from $0.417$ to $0.714$), while recall remains unchanged at $0.833$. This indicates that the baseline model was prone to false positives in this class, which are effectively reduced by CBAM through improved selectivity. In contrast, performance decreases are observed in \textit{Abiotic Stress} (F1 decreases from $0.936$ to $0.826$) and, to a lesser extent, in \textit{Mechanical Stress}. Meanwhile, \textit{Chewing Insect} and \textit{Healthy} remain largely stable. Overall, CBAM shifts EfficientNetB3 towards a more balanced class-wise behavior, increasing macro F1 from $0.806$ to $0.819$, albeit with some degradation in previously well-performing categories.

For InceptionV3, improvements are more uniformly distributed across classes. \textit{Mite Presence} shows the largest gain, with F1 increasing from $0.632$ to $0.833$, primarily driven by a substantial improvement in recall (from $0.500$ to $0.833$), while precision remains stable. \textit{Abiotic Stress} also improves significantly, with F1 increasing from $0.706$ to $0.800$, accompanied by gains in both precision and recall. In contrast, performance decreases are observed in \textit{Chewing Insect} (F1 decreases from $0.667$ to $0.571$) and \textit{Mechanical Stress} (F1 decreases from $0.737$ to $0.667$), mainly due to reduced precision, reflecting a trade-off between sensitivity to minority classes and inter-class discrimination. Despite these effects, overall performance improves, with macro F1 increasing from $0.762$ to $0.791$ and macro recall from $0.701$ to $0.776$.

For EfficientNetB5, the class-wise behavior is more heterogeneous than suggested by aggregate metrics. \textit{Bacterial Spot} achieves perfect classification under CBAM, with precision, recall, and F1-score all equal to $1.000$, while \textit{Abiotic Stress} shows strong improvements, with F1 increasing from $0.884$ to $0.936$. \textit{Mite Presence} also improves substantially in recall (from $0.833$ to $0.917$), leading to an increase in F1 from $0.606$ to $0.733$, despite a reduction in precision. \textit{Chewing Insect} shows a moderate improvement, whereas \textit{Mechanical Stress} exhibits a significant performance drop, with both precision and recall decreasing, suggesting increased confusion with visually similar classes. Nevertheless, macro recall increases from $0.832$ to $0.871$, and macro F1 from $0.837$ to $0.849$.

Across all three architectures, CBAM consistently improves performance on \textit{Mite Presence}, primarily through increased recall in InceptionV3 and EfficientNetB5, and increased precision in EfficientNetB3. EfficientNetB5 with CBAM achieves the best global performance among the three pairs, while InceptionV3 with CBAM shows the largest relative improvement over its own baseline.

\subsection{Transfer Learning for Local Domain Adaptation}

In this second experiment, we tuned the previous models using 4 of the 6 total classes, specifically those corresponding to the local dataset. The selected classes were: \textit{Abiotic Stress}, \textit{Mechanical Stress}, \textit{Healthy}, and \textit{Chewing Insects}.

For this fine-tuning stage, we used images from the local field, as our goal was to analyze how the models adapt to images with different geographical characteristics. The training configuration remained the same as in the previous experiment; however, fine-tuning was limited to 15 epochs to simulate a realistic scenario with constrained data and computational resources. The main objective of this experiment is to improve performance on local data and, consequently, facilitate adaptation to different domains through Transfer Learning.

To this end, we applied the following strategies:

\begin{enumerate}
\item \textbf{Feature Extraction:} We will only train the head in this strategy, aiming to preserve the features extracted from the public domain. This allows us to understand how both domains share common features;
\item \textbf{Fine-tune last:} We fine-tune the last layers of each model, aiming to extract new features from the local domain while retaining previously learned features, such as leaf shape or color;
\item \textbf{Fine-tune all:} We fine-tune the entire model in order to explore whether modifying the weights leads to achieving better performance in the new domain.
\end{enumerate}

We refer to these approaches as Transfer Learning Fine-Tuning strategies (\textit{TL-FT strategies}). We compare their effectiveness to identify the most suitable approach for our models, aiming to obtain robust solutions capable of performing well across both domains without suffering from catastrophic forgetting.

In order to measure how well models generalize, we will use the aforementioned metrics over the local domain, and we split the local dataset into 80/20, using this 80\% of data for fine-tuning. We will also look for $\Delta{F_1}$; it is:

\begin{equation}
\Delta{F_1} = ({F_1})_{AFT} - ({F_1})_{BFT}
\end{equation}

where $AFT$ and $BFT$ refer to after and before fine tuning, respectively. With this, we aim to maximize $\Delta{F_1}$ while improving performance on local data.

Before applying any adaptation strategy, all fourteen source-trained models were evaluated directly on the held-out partition of the local dataset, restricted to the four classes observed in the orchard. This zero-shot evaluation quantifies the cost of deploying models trained on curated public imagery under real acquisition conditions. Performance degraded in 12 of the 14 architectures, with a mean loss of 0.211 macro F1 points, equivalent to a 26.5\% relative reduction with respect to the source domain. The magnitude of the degradation varied widely across architectures: VGG16 suffered the largest drop (0.424 points, 56.5\% relative), followed by MobileNetV3Large (0.372) and EfficientNetB5 (0.340), whereas EfficientNetB3, ResNet101 and InceptionV3 remained essentially stable. Notably, attention-enhanced variants degraded more than their baselines on average (0.260 versus 0.197 points), indicating that CBAM does not by itself confer robustness to distributional shift. The best-performing model in the source domain, EfficientNetB5 + CBAM, retained only 0.548 macro F1 locally, while the strongest zero-shot model was EfficientNetB3, with 0.8108 accuracy and 0.8110 macro F1. These values, reported in the \textit{Before} columns of Table~\ref{tab:fine_tuning_domain_gap_comparison_local}, constitute the baseline against which the adaptation strategies are assessed.

We then explored the three aforementioned strategies by re-training for 15 epochs. The results are summarized in Table~\ref{tab:ft_strategies_f1_acc}. The best strategy metrics (accuracy and F1-score) for each model are highlighted in bold.

\begin{table}[H]
\centering
\caption{F1, Accuracy comparison for each trained model according to the Transfer Learning Fine Tuning strategy (TL-FT) selected.\label{tab:ft_strategies_f1_acc}}
\begin{tabularx}{\textwidth}{lCCCCCC}
\toprule
\multirow{2}{*}{\textbf{Model}} & \multicolumn{2}{c}{\textbf{Feature Extractor}} & \multicolumn{2}{c}{\textbf{Fine-tune Last}} & \multicolumn{2}{c}{\textbf{Fine-tune All}}\\
\cmidrule(lr){2-3} \cmidrule(lr){4-5} \cmidrule(lr){6-7}
& \textbf{F1} & \textbf{Acc} & \textbf{F1} & \textbf{Acc} & \textbf{F1} & \textbf{Acc}\\
\midrule
efficientnetb0 & 0.38 & 0.46 & \textbf{0.61} & \textbf{0.84} & 0.59 & 0.81\\
mobilenetv2 & 0.37 & 0.35 & \textbf{0.90} & \textbf{0.92} & 0.60 & 0.84\\
mobilenetv3large & 0.41 & 0.49 & 0.89 & 0.89 & \textbf{0.92} & \textbf{0.92}\\
efficientnetb3 & 0.45 & 0.57 & 0.85 & 0.86 & \textbf{0.81} & \textbf{0.89}\\
resnet50 & 0.50 & 0.57 & 0.75 & 0.84 & \textbf{0.86} & \textbf{0.86}\\
inceptionv3 & 0.28 & 0.38 & \textbf{0.69} & \textbf{0.73} & 0.72 & 0.81\\
densenet121 & 0.31 & 0.41 & 0.61 & 0.78 & \textbf{0.93} & \textbf{0.92}\\
resnet101 & 0.33 & 0.49 & 0.73 & 0.76 & \textbf{0.86} & \textbf{0.81}\\
vgg16 & 0.26 & 0.41 & \textbf{0.62} & \textbf{0.68} & 0.40 & 0.57\\
vgg19 & 0.33 & 0.30 & \textbf{0.42} & 0.70 & 0.50 & \textbf{0.62}\\
efficientnetb5 & 0.39 & 0.54 & 0.56 & 0.81 & \textbf{0.77} & \textbf{0.84}\\
efficientnetb3 + CBAM & 0.56 & 0.81 & 0.65 & 0.68 & \textbf{0.93} & \textbf{0.95}\\
inceptionv3 + CBAM & 0.57 & 0.81 & \textbf{0.62} & \textbf{0.86} & 0.32 & 0.43\\
efficientnetb5 + CBAM & 0.55 & 0.76 & 0.46 & 0.64 & \textbf{0.82} & \textbf{0.84}\\
\bottomrule
\end{tabularx}
\end{table}

Table~\ref{tab:ft_strategies_f1_acc} shows that fine-tuning all layers (FTA) yields the strongest performance for DenseNet121, ResNet50, ResNet101, EfficientNetB5, and the CBAM-enhanced variants, while MobileNetV2, EfficientNetB0, InceptionV3, VGG16, and InceptionV3 + CBAM achieve their best results under fine-tune-last (FTL). Feature extraction (FE) produces the lowest scores across nearly all evaluated models. Among the evaluated backbones, DenseNet121 and EfficientNetB3 + CBAM achieve the highest overall performance, reaching F1-scores of $0.93$ under full fine-tuning, while ResNet50, ResNet101, and MobileNetV3Large also perform strongly under FTA. InceptionV3 + CBAM, in contrast, shows a marked performance drop under full fine-tuning.

Table~\ref{tab:fine_tuning_domain_gap_comparison_local} reports the performance of each model in the local domain before and after fine-tuning, with the three largest percentage improvements highlighted in bold. MobileNetV3Large improves from an F1-score of $0.3917$ to $0.9218$ ($+135.33$\%), and DenseNet121 improves by $+90.08$\% in F1-score after full fine-tuning. EfficientNetB3 + CBAM achieves the best overall local performance, reaching $0.9459$ accuracy and $0.9297$ F1-score ($+16.67$\% accuracy, $+65.55$\% F1). EfficientNetB5 + CBAM also shows substantial gains ($+50.26$\% F1). In contrast, EfficientNetB0 decreases slightly in both accuracy and F1-score after adaptation, and InceptionV3 and VGG19 show minor F1-score degradations despite improvements in accuracy.

\begin{table}[H]
\centering
\small
\caption{Local performance comparison before and after fine-tuning, including the relative improvement (\%) for the best adaptation strategy.\label{tab:fine_tuning_domain_gap_comparison_local}}
\begin{tabularx}{\textwidth}{llCCCCCC}
\toprule
\multirow{2}{*}{\textbf{Model}} & \multirow{2}{*}{\textbf{FT Method}} & \multicolumn{2}{c}{\textbf{Local Accuracy}} & \multicolumn{2}{c}{\textbf{Local F1}} & \textbf{\% Acc.} & \textbf{\% F1}\\
\cmidrule(lr){3-4} \cmidrule(lr){5-6}
& & Before & After & Before & After & Impro. & Impro.\\
\midrule
mobilenetv2 & FTL & 0.6486 & 0.9189 & 0.6766 & 0.9028 & +41.67\% & +33.43\%\\
efficientnetb0 & FTL & 0.8649 & 0.8378 & 0.6567 & 0.6101 & $-$3.13\% & $-$7.11\%\\
mobilenetv3large & FTA & 0.5135 & \textbf{0.9189} & 0.3917 & \textbf{0.9218} & +78.95\% & +135.33\%\\
densenet121 & FTA & 0.6486 & \textbf{0.9189} & 0.4881 & \textbf{0.9278} & +41.67\% & +90.08\%\\
efficientnetb3 & FTL & 0.8108 & 0.8649 & 0.8110 & 0.8527 & +6.67\% & +5.13\%\\
resnet50 & FTA & 0.7027 & 0.8649 & 0.5293 & 0.8620 & +23.08\% & +62.87\%\\
resnet101 & FTA & 0.7297 & 0.8108 & 0.7848 & 0.8565 & +11.11\% & +9.14\%\\
vgg16 & FTL & 0.3784 & 0.6757 & 0.3260 & 0.6206 & +78.57\% & +90.37\%\\
vgg19 & FTA & 0.5676 & 0.6216 & 0.5180 & 0.4994 & +9.52\% & $-$3.59\%\\
efficientnetb5 & FTA & 0.7297 & 0.8378 & 0.4972 & 0.7708 & +14.81\% & +55.04\%\\
inceptionv3 & FTA & 0.7297 & 0.8108 & 0.7556 & 0.7193 & +11.11\% & $-$4.80\%\\
efficientnetb3 + CBAM & FTA & 0.8108 & \textbf{0.9459} & 0.5616 & \textbf{0.9297} & +16.67\% & +65.55\%\\
efficientnetb5 + CBAM & FTA & 0.7568 & 0.8378 & 0.5476 & 0.8228 & +10.71\% & +50.26\%\\
inceptionv3 + CBAM & FTL & 0.8108 & 0.8649 & 0.5686 & 0.6188 & +6.67\% & +8.81\%\\
\bottomrule
\end{tabularx}
\end{table}

\begin{figure}[H]
\centering
\includegraphics[width=\textwidth]{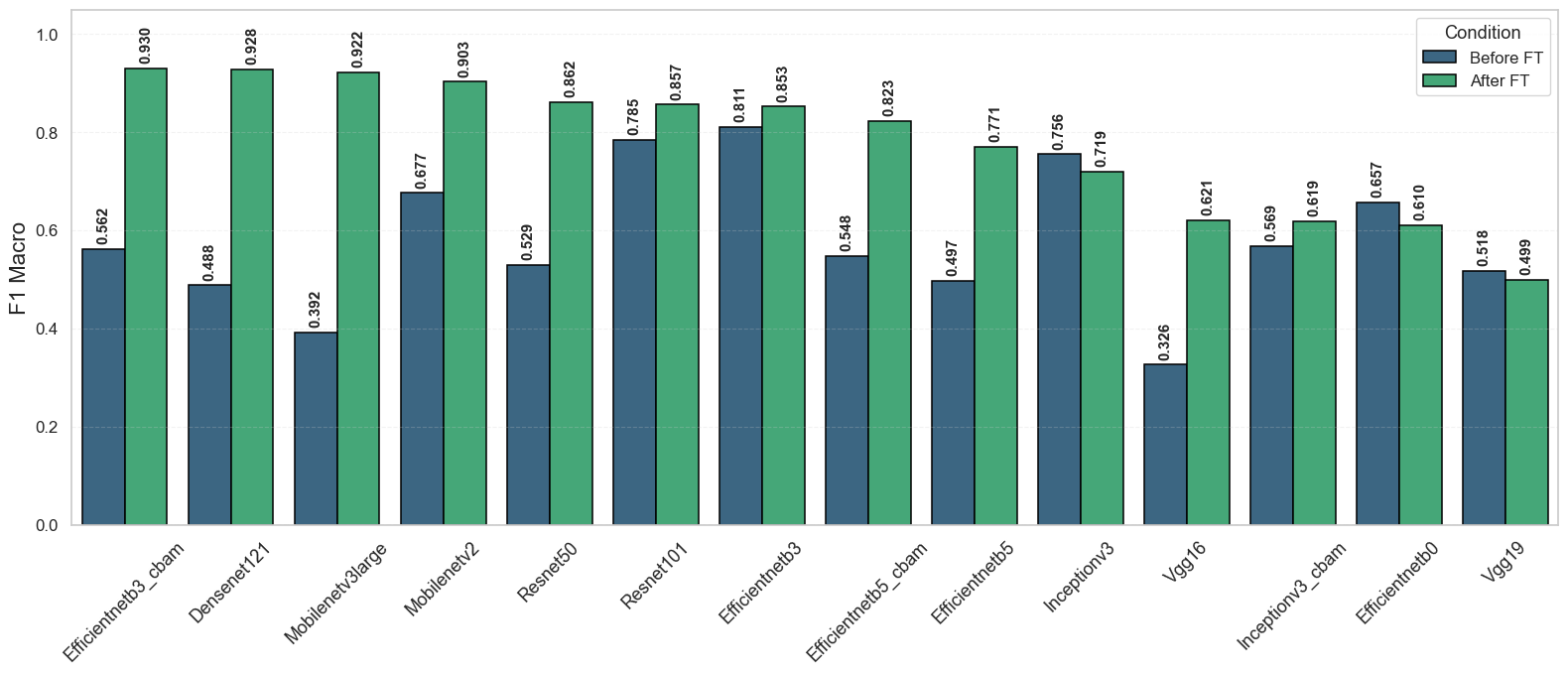}
\caption{F1 (macro) comparison before and after fine-tuning (FT) for each model. The FT strategy plotted is the best one for each model according to Table~\ref{tab:fine_tuning_domain_gap_comparison_local}.\label{fig:deltaF1_models}}
\end{figure}

Figure~\ref{fig:deltaF1_models} shows the change in performance before and after fine-tuning in terms of $\Delta F_1$ (macro-F1 improvement) across all evaluated architectures, including both baseline models and CBAM-augmented variants, using the best fine-tuning strategy for each model as reported in Table~\ref{tab:fine_tuning_domain_gap_comparison_local}. Overall, 11 out of 14 models exhibit a positive $\Delta F_1$. MobileNetV3Large shows the largest gain ($+135$\% in F1-score), followed by EfficientNet-based models and DenseNet121. VGG19 and ResNet101 show the smallest, and in some configurations negative, $\Delta F_1$ values. CBAM-enhanced architectures show consistently positive $\Delta F_1$ values, with the largest gains observed for EfficientNetB3 + CBAM.

\section{Discussion}

The benchmark results highlight three key observations: (i) lightweight and medium-sized architectures can achieve competitive performance without requiring large-scale models; (ii) the EfficientNet family stands out as the best-performing architecture overall; and (iii) the consistent gap between weighted and macro-averaged metrics across all model families indicates that class imbalance remains a limiting factor, particularly for the confusions observed between visually and biologically related classes (\textit{Bacterial Spot} vs.\ \textit{Abiotic Stress}; \textit{Chewing Insect} vs.\ \textit{Mechanical Stress}). This last observation motivates the incorporation of attention mechanisms.

Regarding CBAM, the results indicate that its incorporation does not uniformly improve all performance metrics across all architectures, confirming that the effectiveness of attention mechanisms is architecture-dependent. This contrasts with prior work reporting accuracy improvements across all tested backbones (MobileNet, InceptionV3, and VGG16) after integrating CBAM for tomato leaf disease detection \citep{CBAMTomato}, suggesting that a uniform benefit from CBAM should not be assumed and that its effectiveness must be evaluated on a per-architecture and per-task basis. This heterogeneity becomes particularly relevant in the context of cross-domain adaptation: EfficientNetB3 + CBAM demonstrates the most consistent behavior when transferred to the local dataset, achieving the best balance between source-domain performance and post-adaptation accuracy in the target domain, despite not being the top-performing model in the controlled benchmark setting. At the class level, CBAM consistently improves performance on \textit{Mite Presence}, one of the most challenging minority classes, while precision on dominant classes tends to remain stable or slightly decrease, suggesting that the attention mechanism reduces bias towards majority classes rather than uniformly amplifying predictions. The observed trade-offs in \textit{Mechanical Stress} and \textit{Chewing Insect} are consistent with the visual similarity between these categories and their limited sample sizes.

Concerning Transfer Learning, the results demonstrate that performance on large-scale pretraining benchmarks does not directly translate into robustness in the target domain. Feature extraction consistently produces the weakest results across nearly all evaluated models, confirming that freezing the pretrained backbone is insufficient for this task and that substantial domain adaptation is necessary given the discrepancy between generic features and the visual characteristics of field-collected leaf imagery. Some architectures maintain stable behavior across fine-tuning strategies, whereas others, notably InceptionV3 + CBAM, exhibit considerable sensitivity to the selected configuration, reinforcing that the interaction between attention mechanisms and backbone architecture must be evaluated on a case-by-case basis rather than assumed a priori. This is consistent with the broader literature reporting that supervised domain adaptation strategies for plant disease imagery remain highly setting-dependent \citep{tunio2024advancing,cui2023plant}.

The magnitude of improvement after fine-tuning strongly depends on both backbone and strategy: models with weak baseline performance before adaptation, such as MobileNetV3Large and VGG16, achieve some of the largest relative gains, reinforcing the existence of a substantial domain gap between public-domain pretraining and the local target dataset. This pattern, together with the mostly positive $\Delta F_1$ values observed across models, indicates that Transfer Learning with limited domain-specific data is generally effective for this task, primarily through better adaptation to local visual patterns, although sometimes at a trade-off with performance in the source (public) domain. Deeper or more rigid architectures such as VGG19 and ResNet101 show a higher sensitivity to catastrophic forgetting when fine-tuned on limited local data, reinforcing that model capacity alone does not guarantee better transferability in domain adaptation scenarios. The behaviour of attention-enhanced architectures deserves particular attention, as it is not the one that might be anticipated. Under zero-shot transfer, CBAM variants degraded more than their non-attention counterparts ($0.260$ versus $0.197$ macro F1 points on average), yet after adaptation they reached the highest local scores, with EfficientNetB3~+~CBAM attaining $0.9297$ macro F1. Attention therefore does not appear to make representations invariant to acquisition conditions; rather, the opposite: by recalibrating features channel- and spatially, CBAM plausibly increases the model's reliance on domain-specific cues learned from the source imagery, which is penalised under distributional shift but exploited efficiently once a small number of target samples become available. This reframes attention modules as a mechanism that improves adaptability rather than robustness, a distinction that matters in deployment scenarios where target-domain labels may not be obtainable.

Taken together, these findings address the three gaps identified in the literature: they provide a multi-class benchmark for peach leaf damage under realistic conditions, a systematic evaluation of CBAM in this context, and a direct comparison of fine-tuning strategies for cross-domain adaptation from public to locally collected agricultural data. The results also indicate that findings from controlled, curated benchmarks do not necessarily transfer to field conditions without explicit domain adaptation, a limitation that should be considered when deploying similar models in operational precision agriculture settings.

Several promising directions remain for future research. The synergistic relationship between CBAM and Transfer Learning could be explored in greater depth to further exploit the gains observed in this work, and future studies should evaluate the integration of CBAM into newer architectures, including alternative deployment strategies such as inserting attention modules after every convolutional block, as proposed in the original CBAM framework. Although the incorporation of CBAM increased the number of trainable parameters, its practical impact on inference time and model size was not substantial under the experimental conditions considered here; nevertheless, a more rigorous analysis of computational cost, including inference latency, memory footprint, and energy consumption on resource-constrained hardware, remains necessary before deploying CBAM-enhanced models on edge and mobile devices for real-time in-field diagnosis. Finally, extending this study to a broader range of crops and plant diseases would help determine whether the conclusions obtained from peach leaf imagery generalize to other agricultural scenarios and fruit species.

\section{Conclusions}

This work addressed peach leaf damage classification as a transferability problem rather than a purely architectural one. Eleven convolutional backbones and three CBAM-enhanced variants were benchmarked on a manually annotated public dataset of 1366 leaves across six damage categories, where CBAM-EfficientNetB5 achieved the best source-domain performance ($93.3$\% accuracy, $0.849$ macro F1). The same models were then evaluated on an independently acquired dataset of 180 images collected in a commercial orchard and released publicly, which served as an unseen target domain. Twelve of the fourteen architectures degraded under this shift, losing $0.211$ macro F1 points on average, and the ranking obtained in the controlled benchmark did not predict local behaviour. Supervised fine-tuning recovered most of this loss: feature extraction proved insufficient in nearly all cases, whereas full or partial fine-tuning with fewer than 150 target images raised EfficientNetB3~+~CBAM to $0.9459$ accuracy and $0.9297$ macro F1 on the local domain.

Two findings carry practical implications for precision agriculture deployments. First, performance reported on curated public benchmarks systematically overestimates field behaviour, and the magnitude of this overestimation is architecture-dependent, so model selection should be validated on target-domain imagery rather than inherited from published rankings. Second, attention mechanisms improved minority-class detection and post-adaptation accuracy but did not reduce the degradation observed under zero-shot transfer, indicating that CBAM enhances adaptability rather than robustness. Practitioners should therefore budget for a small annotated sample from the deployment site instead of expecting attention-based architectures to generalise without adaptation.

\section*{Author Contributions}
Writing---review and editing, writing---original draft, software, methodology, investigation, validation, A.C.-R.; writing---review and editing, writing---original draft, data curation, software, validation, methodology, investigation, M.A.G.-I.; writing---review and editing, resources, conceptualization, M.F.G.-C.; resources, P.N.T.; resources, J.S.R.-A.; writing---review and editing, conceptualization, resources, M.A.Z.-I.; writing---review and editing, J.A.M.N.; writing---review and editing, supervision, funding acquisition, conceptualization, A.F.S. All authors have read and agreed to this version of the manuscript.

\section*{Funding}
This work has been funded by the project OSIRIS (TSI-100921-2023-1) under the \textit{C\'atedras ENIA}, promoted by the Ministry for Digital Transformation and Public Function and the Secretar\'ia de Estado de Digitalizaci\'on e Inteligencia Artificial, and co-funded by the European Union--NextGenerationEU, through the PRTR.

\section*{Data Availability Statement}
The local peach leaf dataset collected in this study (180 images across four classes, acquired in a commercial orchard in Jumilla, Murcia) is openly available on Zenodo at \url{https://doi.org/10.5281/zenodo.20446065}. The benchmark dataset was derived from publicly available sources \citep{PlantDocDataset,DatasetUsed} and is therefore subject to their original licensing terms. The source code for data preprocessing, model training, the CBAM implementation, and the transfer-learning experiments is publicly available at \url{https://github.com/adricanovas/peach-leaf-cbam}.

\section*{Acknowledgments}
During the preparation of this work, the authors used Google Gemini and DeepL Write to check the grammar of the manuscript and improve the clarity of the more technical sections. After using these tools, the authors reviewed and edited the content as necessary and take full responsibility for the content of the article.

\section*{Conflicts of Interest}
The authors declare that they have no known competing financial interests or personal relationships that could have appeared to influence the work reported in this paper.

\section*{Abbreviations}
The following abbreviations are used in this manuscript:

\medskip
\noindent
\begin{tabular}{@{}ll}
CBAM & Convolutional Block Attention Module\\
CNN & Convolutional Neural Network\\
COCO & Common Objects in Context\\
CVAT & Computer Vision Annotation Tool\\
FE & Feature Extraction\\
FTA & Fine-Tune All\\
FTL & Fine-Tune Last\\
GAP & Global Average Pooling\\
kNN & k-Nearest Neighbors\\
MLP & Multilayer Perceptron\\
ReLU & Rectified Linear Unit\\
SE & Squeeze-and-Excitation\\
SVM & Support Vector Machine\\
TL-FT & Transfer Learning Fine-Tuning
\end{tabular}



\begin{thebibliography}{99}
\small

\bibitem{murugavalli2025plant}
Murugavalli, S.; Gopi, R. Plant leaf disease detection using vision transformers for precision agriculture. \textit{Sci. Rep.} \textbf{2025}, \textit{15}, 22361.

\bibitem{zhao2025review}
Zhao, J.; Xu, L.; Ma, Z.; Li, J.; Wang, X.; Liu, Y.; Du, X. A review of plant leaf disease identification by deep learning algorithms. \textit{Front. Plant Sci.} \textbf{2025}, \textit{16}, 1637241.

\bibitem{upadhyay2025deep}
Upadhyay, A.; Chandel, N.S.; Singh, K.P.; Chakraborty, S.K.; Nandede, B.M.; Kumar, M.; Subeesh, A.; Upendar, K.; Salem, A.; Elbeltagi, A. Deep learning and computer vision in plant disease detection: a comprehensive review of techniques, models, and trends in precision agriculture. \textit{Artif. Intell. Rev.} \textbf{2025}, \textit{58}, 92.

\bibitem{krishna2025plant}
Krishna, M.S.; Machado, P.; Otuka, R.I.; Yahaya, S.W.; Neves dos Santos, F.; Ihianle, I.K. Plant Leaf Disease Detection Using Deep Learning: A Multi-Dataset Approach. \textit{J} \textbf{2025}, \textit{8}, 4.

\bibitem{hu2025domain}
Hu, X.; Chen, S.; Zhang, D. Domain Adaptation in Agricultural Image Analysis: A Comprehensive Review from Shallow Models to Deep Learning. \textit{arXiv} \textbf{2025}, arXiv:2506.05972.

\bibitem{shafik2025deep}
Shafik, W.; Tufail, A.; De Silva, L.C.; Haji Mohd Apong, R.A.A.; Kim, K.H. Deep learning technique for plant disease classification and pest detection and model explainability elevating agricultural sustainability. \textit{BMC Plant Biol.} \textbf{2025}, \textit{25}, 1491.

\bibitem{PeachCNN}
Yadav, S.; Sengar, N.; Singh, A.; Singh, A.; Dutta, M.K. Identification of disease using deep learning and evaluation of bacteriosis in peach leaf. \textit{Ecol. Inform.} \textbf{2021}, \textit{61}, 101247.

\bibitem{CNN_CAE}
Bedi, P.; Gole, P. Plant disease detection using hybrid model based on convolutional autoencoder and convolutional neural network. \textit{Artif. Intell. Agric.} \textbf{2021}, \textit{5}, 90--101.

\bibitem{DatasetOriginalArticle}
Farman, H.; Ahmad, J.; Jan, B.; Shahzad, Y.; Abdullah, M.; Ullah, A. EfficientNet-Based Robust Recognition of Peach Plant Diseases in Field Images. \textit{Comput. Mater. Contin.} \textbf{2021}, \textit{71}, 2073--2089.

\bibitem{alosaimi2021peachnet}
Alosaimi, W.; Alyami, H.; Uddin, M.I. PeachNet: Peach Diseases Detection for Automatic Harvesting. \textit{Comput. Mater. Contin.} \textbf{2021}, \textit{67}.

\bibitem{PeachEfB7}
Ranga, S.; Sheoran, S.K.; Singh, G. Leaf Diseases Detection in Peach Using EfficientNet. In Proceedings of the International Conference on Electronic Governance with Emerging Technologies, 2024; pp. 109--121.

\bibitem{benchmarktomato}
Brahimi, M.; Arsenovic, M.; Laraba, S.; Sladojevic, S.; Boukhalfa, K.; Moussaoui, A. Deep Learning for Plant Diseases: Detection and Saliency Map Visualisation. In \textit{Human and Machine Learning: Visible}; Springer International Publishing: Cham, Switzerland, 2018; pp. 93--117.

\bibitem{cropsCNN}
Monigari, V. Plant Leaf Disease Prediction. \textit{Int. J. Res. Appl. Sci. Eng. Technol.} \textbf{2021}, \textit{9}, 1295--1305.

\bibitem{RiceCNN}
Krishnamoorthy, N.; Narasimha Prasad, L.V.; Pavan Kumar, C.S.; Subedi, B.; Abraha, H.B.; Sathishkumar, V.E. Rice leaf diseases prediction using deep neural networks with transfer learning. \textit{Environ. Res.} \textbf{2021}, \textit{198}, 111275.

\bibitem{tomatobenchmark}
Das, A.; Pathan, F.; Jim, J.R.; Kabir, M.M.; Mridha, M. Deep learning-based classification, detection, and segmentation of tomato leaf diseases: A state-of-the-art review. \textit{Artif. Intell. Agric.} \textbf{2025}, \textit{15}, 192--220.

\bibitem{DLandML}
Sujatha, R.; Krishnan, S.; Chatterjee, J.M.; Gandomi, A.H. Advancing plant leaf disease detection integrating machine learning and deep learning. \textit{Sci. Rep.} \textbf{2025}, \textit{15}, 11552.

\bibitem{DenseNetlite}
Nobel, S.N.; Afroj, M.; Kabir, M.M.; Mridha, M. Development of a cutting-edge ensemble pipeline for rapid and accurate diagnosis of plant leaf diseases. \textit{Artif. Intell. Agric.} \textbf{2024}, \textit{14}, 56--72.

\bibitem{SEpaper}
Hu, J.; Shen, L.; Albanie, S.; Sun, G.; Wu, E. Squeeze-and-Excitation Networks. \textit{arXiv} \textbf{2019}, arXiv:1709.01507.

\bibitem{CBAMpaper}
Woo, S.; Park, J.; Lee, J.; Kweon, I.S. CBAM: Convolutional Block Attention Module. \textit{arXiv} \textbf{2018}, arXiv:1807.06521.

\bibitem{MobileNetv2CBAM}
Rakha, M.; Sulistiyo, M.D.; Nasien, D.; Ridha, M. A Combined MobileNetV2 and CBAM Model to Improve Classifying the Breast Cancer Ultrasound Images. \textit{J. Appl. Eng. Technol. Sci.} \textbf{2024}, \textit{6}, 561--578.

\bibitem{EfNetCBAM}
Ul Amin, S.; Sibtain Abbas, M.; Kim, B.; Jung, Y.; Seo, S. Enhanced Anomaly Detection in Pandemic Surveillance Videos: An Attention Approach With EfficientNet-B0 and CBAM Integration. \textit{IEEE Access} \textbf{2024}, \textit{12}, 162697--162712.

\bibitem{CBAMTomato}
G\"ok, A.E.; Yurdakul, M.; Tasdemir, S. CBAM Enhanced Deep Learning Models for Plant Disease Detection. In Proceedings of the 2nd International Conference on Trends in Advanced Research (ICTAR 2024), T\"urkiye, 2024.

\bibitem{TLtomatoandgrave}
Paymode, A.S.; Malode, V.B. Transfer Learning for Multi-Crop Leaf Disease Image Classification using Convolutional Neural Network VGG. \textit{Artif. Intell. Agric.} \textbf{2022}, \textit{6}, 23--33.

\bibitem{DeepRice}
Ritharson, P.I.; Raimond, K.; Mary, X.A.; Robert, J.E.; Andrew, J. DeepRice: A deep learning and deep feature based classification of Rice leaf disease subtypes. \textit{Artif. Intell. Agric.} \textbf{2024}, \textit{11}, 34--49.

\bibitem{DomainAdaptation}
Mansour, Y. Learning and Domain Adaptation. \textit{Lect. Notes Comput. Sci.} \textbf{2009}, \textit{5808}, 4--6.

\bibitem{TransferLearningvsDomainAdaptation}
Kouw, W.M.; Loog, M. An introduction to domain adaptation and transfer learning. \textit{arXiv} \textbf{2019}, arXiv:1812.11806.

\bibitem{TransferLearning}
Zhuang, F.; Qi, Z.; Duan, K.; Xi, D.; Zhu, Y.; Zhu, H.; Xiong, H.; He, Q. A Comprehensive Survey on Transfer Learning. \textit{arXiv} \textbf{2020}, arXiv:1911.02685.

\bibitem{SDA}
Goodman, S.; Greenspan, H.; Goldberger, J. Supervised Domain Adaptation by transferring both the parameter set and its gradient. \textit{Neurocomputing} \textbf{2023}, \textit{560}, 126828.

\bibitem{tunio2024advancing}
Tunio, M.H.; Li, J.P.; Zeng, X.; Ahmed, A.; Shah, S.A.; Shaikh, H.U.; Mallah, G.A.; Yahya, I.A. Advancing plant disease classification: A robust and generalized approach with transformer-fused convolution and Wasserstein domain adaptation. \textit{Comput. Electron. Agric.} \textbf{2024}, \textit{227}, 109574.

\bibitem{cui2023plant}
Cui, Z.; Li, K.; Kang, C.; Wu, Y.; Li, T.; Li, M. Plant and disease recognition based on PMF pipeline domain adaptation method: Using bark images as meta-dataset. \textit{Plants} \textbf{2023}, \textit{12}, 3280.

\bibitem{PlantDocDataset}
Singh, D.; Jain, N.; Jain, P.; Kayal, P.; Kumawat, S.; Batra, N. PlantDoc: A Dataset for Visual Plant Disease Detection. In Proceedings of the 7th ACM IKDD CoDS and 25th COMAD, New York, NY, USA, 2020; pp. 249--253.

\bibitem{DatasetUsed}
Ahmad, J.; Farman, H. Dataset: Efficientnet-based robust recognition of peach plant diseases in field images. Mendeley Data, V1, 2024. Available online: \url{https://data.mendeley.com/datasets/3pmj85snvw/1} (accessed on 8 September 2026).

\bibitem{CVAT}
Sekachev, B.; Manovich, N.; Zhiltsov, M.; Zhavoronkov, A.; Kalinin, D.; Hoff, B.; Osmanov, T.; Kruchinin, D.; Zankevich, A.; Sidnev, D.; et al. opencv/cvat: v1.1.0. Zenodo, 2020. Available online: \url{https://doi.org/10.5281/zenodo.4009388} (accessed on 8 September 2026).

\bibitem{luo2022identification}
Luo, M.; Meng, F.Z.; Tan, Q.; Zhou, Y.; Chaisiri, C.; Fan, F.; Yin, W.X.; Luo, C.X. Identification, genetic diversity, and chemical control of Xanthomonas arboricola pv. pruni in China. \textit{Plant Dis.} \textbf{2022}, \textit{106}, 2415--2423.

\bibitem{garita2018xanthomonas}
Garita-Cambronero, J.; Palacio-Bielsa, A.; Cubero, J. Xanthomonas arboricola pv. pruni, causal agent of bacterial spot of stone fruits and almond: its genomic and phenotypic characteristics in the X. arboricola species context. \textit{Mol. Plant Pathol.} \textbf{2018}, \textit{19}, 2053--2065.

\bibitem{bacterial_treatment}
Sakata, N.; Ishiga, Y. Advancing Sustainable Management of Bacterial Spot of Peaches: Insights into Xanthomonas arboricola pv. pruni Pathogenicity and Control Strategies. \textit{Bacteria} \textbf{2025}, \textit{4}.

\bibitem{abiotic_visual_detection}
Vollenweider, P.; G\"unthardt-Goerg, M.S. Diagnosis of abiotic and biotic stress factors using the visible symptoms in foliage. \textit{Environ. Pollut.} \textbf{2005}, \textit{137}, 455--465.

\bibitem{joshi2023management}
Joshi, N.K.; Phan, N.T.; Biddinger, D.J. Management of Panonychus ulmi with various miticides and insecticides and their toxicity to predatory mites conserved for biological mite control in eastern US Apple Orchards. \textit{Insects} \textbf{2023}, \textit{14}, 228.

\bibitem{mite_underside}
Sudo, M.; Osakabe, M. Do plant mites commonly prefer the underside of leaves? \textit{Exp. Appl. Acarol.} \textbf{2011}, \textit{55}, 25--38.

\bibitem{Mahendiran2022}
Mahendiran, G.; Lal, S.; Sharma, O.C. Pests and Their Management on Temperate Fruits. In \textit{Trends in Horticultural Entomology}; Springer Nature Singapore: Singapore, 2022; pp. 891--941.

\bibitem{medda2022influence}
Medda, S.; Fadda, A.; Mulas, M. Influence of climate change on metabolism and biological characteristics in perennial woody fruit crops in the Mediterranean environment. \textit{Horticulturae} \textbf{2022}, \textit{8}, 273.

\bibitem{keras}
Chollet, F. Keras, 2015. Available online: \url{https://keras.io} (accessed on 8 September 2026).

\bibitem{ImageNet}
Deng, J.; Dong, W.; Socher, R.; Li, L.J.; Li, K.; Fei-Fei, L. ImageNet: A large-scale hierarchical image database. In Proceedings of the 2009 IEEE Conference on Computer Vision and Pattern Recognition, 2009; pp. 248--255.

\bibitem{Albumentations}
Buslaev, A.; Iglovikov, V.I.; Khvedchenya, E.; Parinov, A.; Druzhinin, M.; Kalinin, A.A. Albumentations: Fast and Flexible Image Augmentations. \textit{Information} \textbf{2020}, \textit{11}, 125.

\bibitem{MobileNetpaper}
Howard, A.G.; Zhu, M.; Chen, B.; Kalenichenko, D.; Wang, W.; Weyand, T.; Andreetto, M.; Adam, H. MobileNets: Efficient Convolutional Neural Networks for Mobile Vision Applications. \textit{arXiv} \textbf{2017}, arXiv:1704.04861.

\bibitem{MobileNetv2}
Sandler, M.; Howard, A.; Zhu, M.; Zhmoginov, A.; Chen, L. MobileNetV2: Inverted Residuals and Linear Bottlenecks. \textit{arXiv} \textbf{2019}, arXiv:1801.04381.

\bibitem{MobileNetv3}
Howard, A.; Sandler, M.; Chu, G.; Chen, L.; Chen, B.; Tan, M.; Wang, W.; Zhu, Y.; Pang, R.; Vasudevan, V.; Le, Q.V.; Adam, H. Searching for MobileNetV3. \textit{arXiv} \textbf{2019}, arXiv:1905.02244.

\bibitem{EfficientNet}
Tan, M.; Le, Q.V. EfficientNet: Rethinking Model Scaling for Convolutional Neural Networks. \textit{arXiv} \textbf{2020}, arXiv:1905.11946.

\bibitem{ResNet}
He, K.; Zhang, X.; Ren, S.; Sun, J. Deep Residual Learning for Image Recognition. \textit{arXiv} \textbf{2015}, arXiv:1512.03385.

\bibitem{VGG}
Simonyan, K.; Zisserman, A. Very Deep Convolutional Networks for Large-Scale Image Recognition. \textit{arXiv} \textbf{2015}, arXiv:1409.1556.

\bibitem{InceptionV3}
Szegedy, C.; Vanhoucke, V.; Ioffe, S.; Shlens, J.; Wojna, Z. Rethinking the Inception Architecture for Computer Vision. \textit{arXiv} \textbf{2015}, arXiv:1512.00567.

\bibitem{DenseNet}
Huang, G.; Liu, Z.; van der Maaten, L.; Weinberger, K.Q. Densely Connected Convolutional Networks. \textit{arXiv} \textbf{2018}, arXiv:1608.06993.

\bibitem{Adam}
Kingma, D.P.; Ba, J. Adam: A Method for Stochastic Optimization. \textit{arXiv} \textbf{2017}, arXiv:1412.6980.

\bibitem{Tensorflow}
Abadi, M.; Agarwal, A.; Barham, P.; Brevdo, E.; Chen, Z.; Citro, C.; Corrado, G.S.; Davis, A.; Dean, J.; Devin, M.; et al. TensorFlow: Large-Scale Machine Learning on Heterogeneous Systems, 2015. Available online: \url{https://www.tensorflow.org/} (accessed on 8 September 2026).

\bibitem{Pytorch}
Paszke, A.; Gross, S.; Massa, F.; Lerer, A.; Bradbury, J.; Chanan, G.; Killeen, T.; Lin, Z.; Gimelshein, N.; Antiga, L.; et al. PyTorch: An Imperative Style, High-Performance Deep Learning Library. \textit{arXiv} \textbf{2019}, arXiv:1912.01703.

\end{thebibliography}
\end{document}